\documentclass{article}

\usepackage{graphicx} 
\usepackage{amssymb}
\usepackage{amstext}
\usepackage{amsmath}
\usepackage{amsthm}
\usepackage{multicol}
\usepackage{booktabs}
\usepackage{url}

\newcommand{\eatdot}[1]{\ifx.#1\else#1\fi}

\begin{document}

\title{Is Chaotic Dynamics Enhancing Particle Swarm Optimization?\\ An Anytime Performance Assessment}

\author{Hendrik Richter\\
HTWK Leipzig University of Applied Sciences, Faculty of Engineering\\ Leipzig, Germany, Email: hendrik.richter@htwk-leipzig.de}

\maketitle

\section*{Abstract}
The search dynamics of particle swarm optimization (PSO) can be driven by random or chaotic values. We study if there are performance differences between random and chaos if distributional equivalence of the used sequences is observed. Therefore, we consider an anytime performance assessment which quantifies the fraction of runs reaching a given target value within a specified numerical budget. Furthermore, we connect quantifiers of the chaotic dynamics to performance metrics of the anytime analysis. The results show that once the underlying distributions are controlled, deterministic chaotic sequences yield no general or consistent advantage over random sequences. Performance differences arise most prominently from the underlying distribution, but also from search-space dimensions, benchmark targets, and numerical budgets.  We also find potential budget-dependent advantages under specific parametrization of chaotic maps and distribution settings.

\section{Introduction}
The ability of chaotic systems to generate random-like values has led to claims that chaos may generally improve optimization performance of metaheuristic algorithms, if compared to conventional random number generation~\cite{ala09,chen18,xu18,tian19,zel22,liman25,zhao25}. However, opposing (or at least ambivalent) findings have also been reported~\cite{yan17,gag21,noer23,rich24}. 

Central to understanding how and why chaos may enhance performance is the question of how sequences of random and chaotic values may actually differ. Such differences can be characterized either statistically by considering a large ensemble of values and analyzing the resulting distribution, or dynamically by examining shorter sequences and focusing on the temporal dependencies between successive values. For independently generated random values, successive values are mutually independent such that their statistical properties are characterized solely by their probability distribution.  In contrast, successive chaotic values are deterministically related through a  nonlinear mapping.  In other words, for chaos generated by different nonlinear maps (or by different parameters of the same map),  distribution  \emph{and} temporal interdependency may vary. Thus, for a fair comparison of random and chaos, we need to separate the distributional effects (which likelihood have certain values) from the temporal-dependence effects (how are successive values correlated).

Recently, results on particle swarm optimization (PSO) have taken into account the separation of distributional and temporal effects of chaos, and argued that 
distributions have the main effect on performance~\cite{noer23,rich24}. However, these results also showed that significantly smaller but clearly discernible performance differences between random and chaos remain even if the underlying distribution is the same.  Here, we look at these differences (and their potential causes) through the lens of anytime performance assessment~\cite{hans22,ye22,lo24}. As in the studies mentioned~\cite{noer23,rich24}, we study PSO as its search dynamics is directly driven by random-like quantities. Thus, the effects of using random or chaotic values can be studied in a particularly straightforward and unambiguous way. The main contributions of this study are:
\begin{itemize}
  \item arranging an experimental setup with parametrizable chaotic maps, which  consistently have values with the same distribution, but varying chaotic dynamics properties. This allows us to isolate dynamics properties of chaos from distribution effects,  
  \item giving (as a novelty of random vs. chaos comparisons) an anytime performance analysis which accounts for reaching a given target within a specified numerical budget,  
  \item linking performance metrics obtained by anytime analysis to quantities measuring chaotic dynamics. In particular, we study Lyapunov exponents~\cite{ott93}, but also mutual information~\cite{fras86} and permutation entropy~\cite{ban02}, which can be calculated for both random and chaotic sequences, and
  \item showing that the results seriously challenge the assumption that chaos generally improves PSO performance, irrespective of distributional and temporal properties. These doubts become even more pronounced when considering anytime performance over a numerical budget.  
\end{itemize}
The paper is structured as follows. Sec.~\ref{sec:rand_chao}
briefly recalls relevant concepts about random and chaos, and describes how for chaotic values distributional equivalence can be set up and temporal differences can be quantified. Sec.~\ref{sec:setup} details the experimental setup used, namely the PSO algorithm,  the BBOB benchmark suite, and the anytime performance metrics. Results are presented and discussed in Sec.~\ref{sec:results}, while the final section provides concluding remarks.

\section{Random vs. Chaos} \label{sec:rand_chao}

  In numerical experiments, the randomness needed in metaheuristic algorithms is typically implemented by employing pseudo-random number generators (PRNGs) \cite{lec12}. A key characteristic of PRNGs is their ability to generate values that follow a prescribed probability distribution while ensuring statistical independence between successive values. 
An alternative means of generating random-like values involves using time series produced by chaotic maps. Some chaotic maps produce sequences whose values are distributed according to distributions commonly used in metaheuristics, such as the uniform distribution and the arcsine distribution (also known as U-shaped distribution). Contrary to PRNGs, successive values of chaotic time series are not statistically independent, in fact they are deterministic with respect to the nonlinearity of the map.

A central aspect of the analysis is that we fix the distributions from which the values are drawn. Furthermore, we employ chaotic maps in which the distribution of generated values remains invariant under parameter changes, while quantitative measures of chaos can be tuned.
Next, it is shown how such an arrangement can be established.

\subsection{Setting up Distributional Equivalence}

In the following, we consider continuous (probability) distributions of a variable $z\in\mathbb{R}$ defined in the 
interval $[0,1]$, taking as examples the uniform distribution $\varrho(z)=1$
and the arcsine distribution
$\varrho(z)=\pi^{-1} \left(z(1-z)\right)^{-\frac{1}{2}}$. 
Note that both distributions are instances of a Beta distribution $\mathcal{B}(z,\alpha,\beta)=z^{\alpha-1}(1-z)^{\beta-1} \cdot \mathbb{B}(\alpha,\beta)^{-1}$ with the Beta function $\mathbb{B}(\alpha,\beta)$. For $\alpha=\beta=1$, we get $\mathbb{B}(1,1)=1$, and a uniform distribution, while for $\alpha=\beta=0.5$, $\mathbb{B}(0.5,0.5)=\pi$, and we have an arcsine distribution.

For chaotic maps, the distribution of values is described by the natural invariant density~\cite{ott93,diak96}. Suppose that the natural invariant density matches the probability density function associated with the distribution. Then successive values of the map are 
statistically equivalent to realizations of a random variable distributed according to the same function. This, for instance, applies for numerical values $z(k)$ produced by a PRNG (with $k$ denoting the time variable, $k=0,1,2,\ldots$, indicating the order in which the values are generated). We next consider chaotic maps that retain their invariant densities under parameter variation, while their chaotic dynamics changes.

An example of a map with consistently arcsine-distributed values is the Schröder map~\cite{lue08} 
\begin{equation}z(k+1)=\sin^2\left(b \cdot \arcsin{\sqrt{x(k)}} \right) \label{eq:schr}\end{equation}
which also uses the time variable $k$ and whose chaotic characteristics can be tuned by $b \in \mathbb{N}$, $b \geq 2$. 
It should be noted that 
the frequently studied Logistic map
\begin{equation}z(k+1)=rz(k)(1-z(k)) \label{eq:logist} \end{equation}
for $r=4$ also generates values observing an arcsine distribution. Moreover, the map \eqref{eq:logist} with $r=4$ is identical to the Schröder map  \eqref{eq:schr} with $b=2$, see the supplemental material, and thus included in the numerical experiments. 
Unfortunately, the Logistic map does not allow us to tune quantitative measures of chaos while simultaneously preserving distributional equivalence. If we decrease $r$ to values $r<4$, we may still have chaos with changed chaotic characteristics, but we also abandon the arcsine distribution. 

\begin{figure}

\includegraphics[trim = 9mm 3mm 10mm 8mm,clip, width=6.2cm, height=4.5cm]{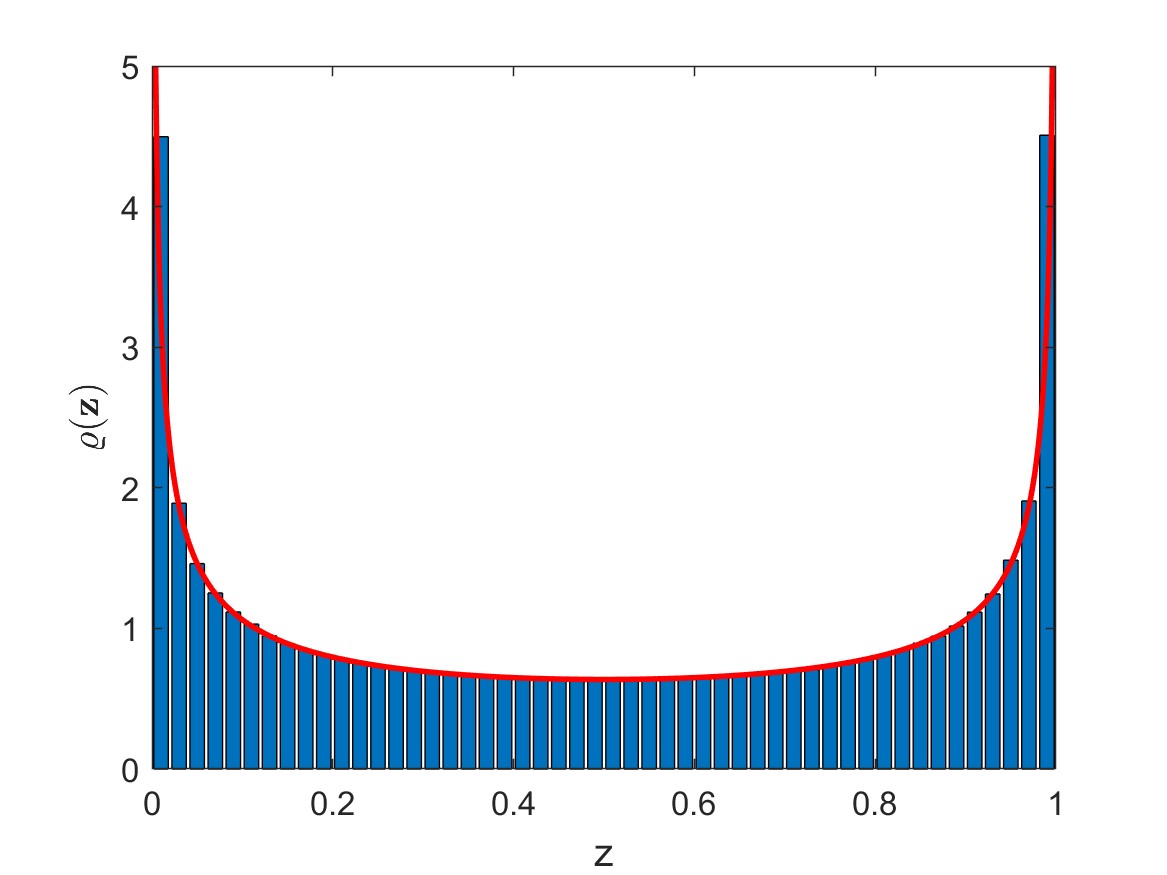}
\includegraphics[trim = 9mm 3mm 10mm 8mm,clip, width=6.2cm, height=4.5cm]{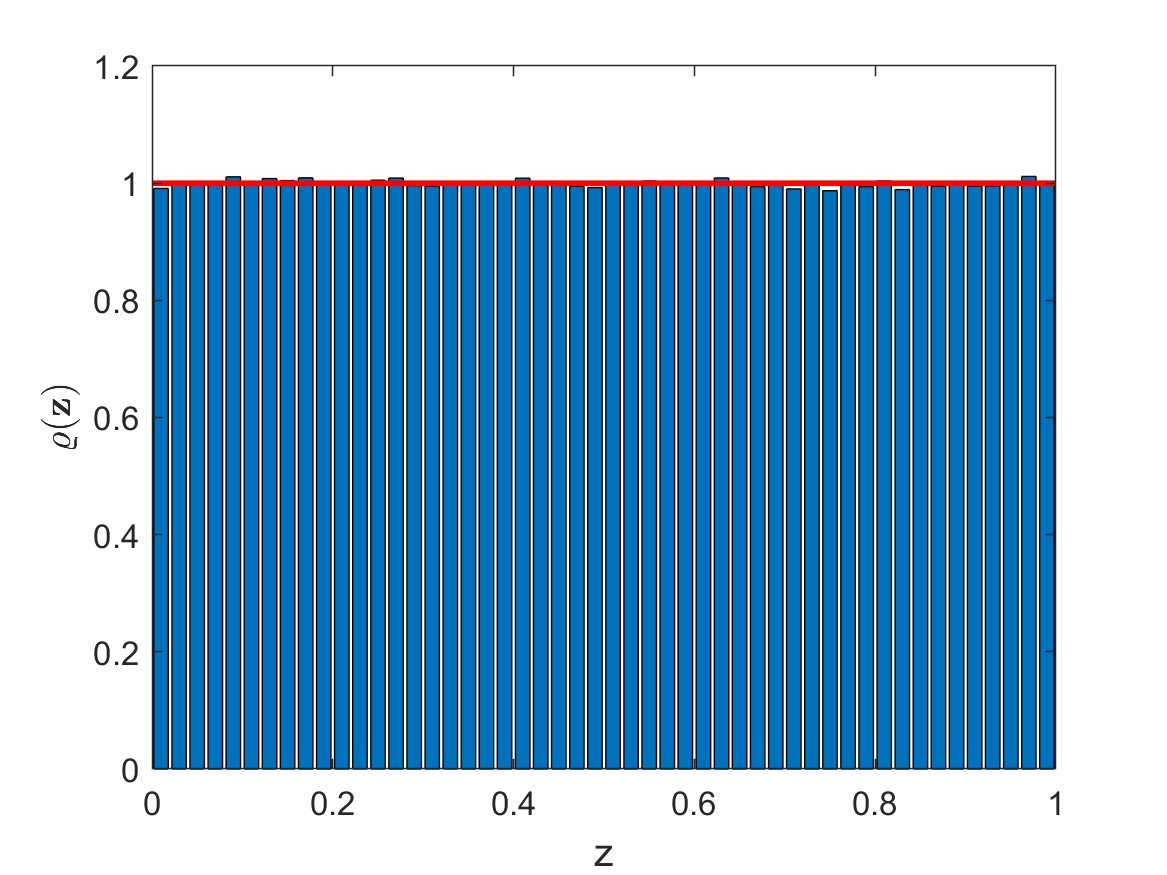}

(a) Arcsine: Eq. \eqref{eq:schr}, $b=2$  \hspace{1cm} (b) Uniform:  Eq. \eqref{eq:atent}, $a=0.65$

\caption{Distributions and histograms of values considered in the experiments. (a) Arcsine: Schröder map,  Eq. \eqref{eq:schr}, distribution
$\varrho(z)=\pi^{-1} \left(z(1-z)\right)^{-\frac{1}{2}}$.  (b) Uniform: asymmetric tent map,   Eq. \eqref{eq:atent}, distribution 
$\varrho(z)=1$. }
\label{fig_dist}
\end{figure}

An example of a parametrizable chaotic map with a uniform density is the
asymmetric tent map (also skew tent map)
\begin{equation}z(k+1)=\min{\left(\frac{z(k)}{a},\frac{1-z(k)}{1-a}\right)}.   \label{eq:atent} \end{equation}
It can be tuned by a parameter $a \in \mathbb{R}$ and maintains uniformly distributed values for $0<a<1$, while  quantitative measures of chaos vary. The map \eqref{eq:atent} is equal to the standard tent map for $a=0.5$~\cite{ott93}. Fig. \ref{fig_dist} shows the distributions of values with arcsine and uniform density as histograms obtained by the Schröder map, Eq. \eqref{eq:schr}, with $b=2$ and the asymmetric tent map, Eq. \eqref{eq:atent}, with $a=0.65$, together with the distribution curves.

\subsection{Quantifying Temporal Differences} \label{sec:temp_diff}

\begin{figure}

\includegraphics[trim = 9mm 3mm 10mm 8mm,clip, width=6.2cm, height=4.5cm]{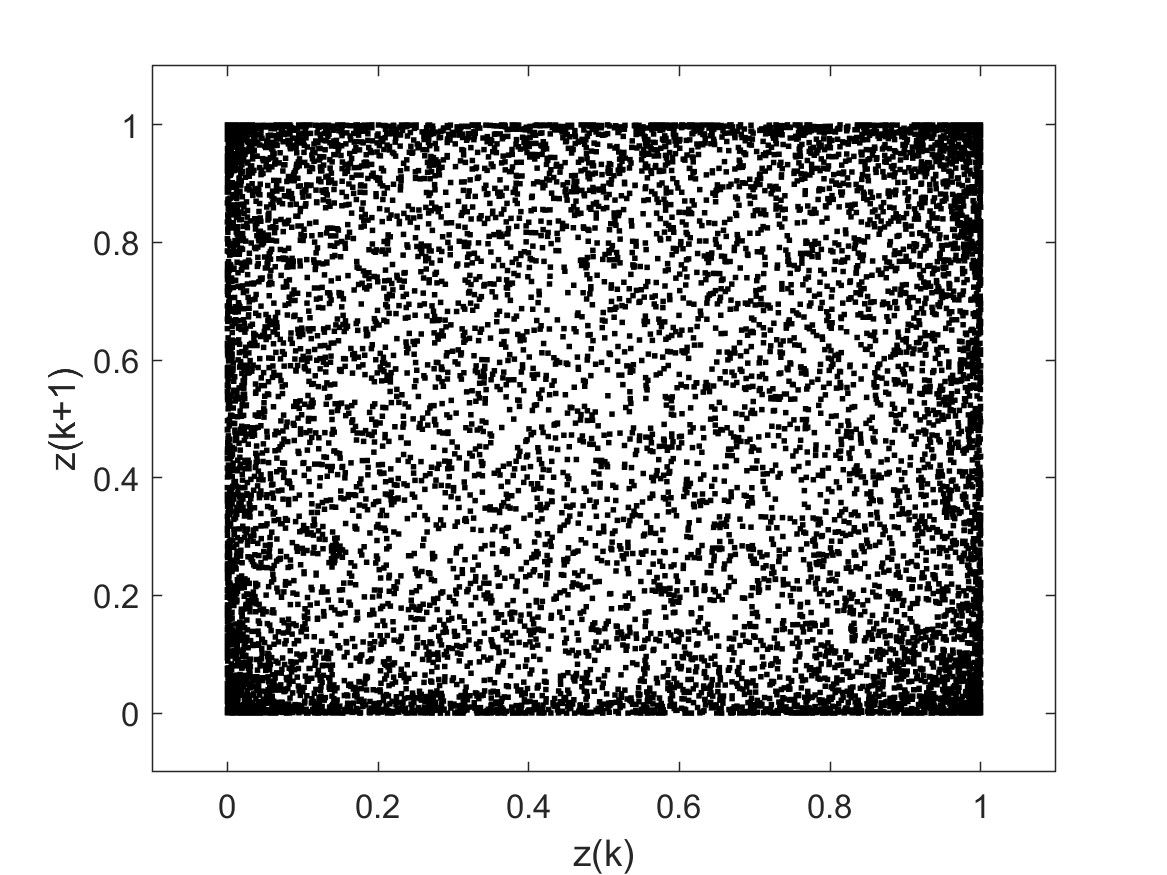}
\includegraphics[trim = 9mm 3mm 10mm 8mm,clip, width=6.2cm, height=4.5cm]{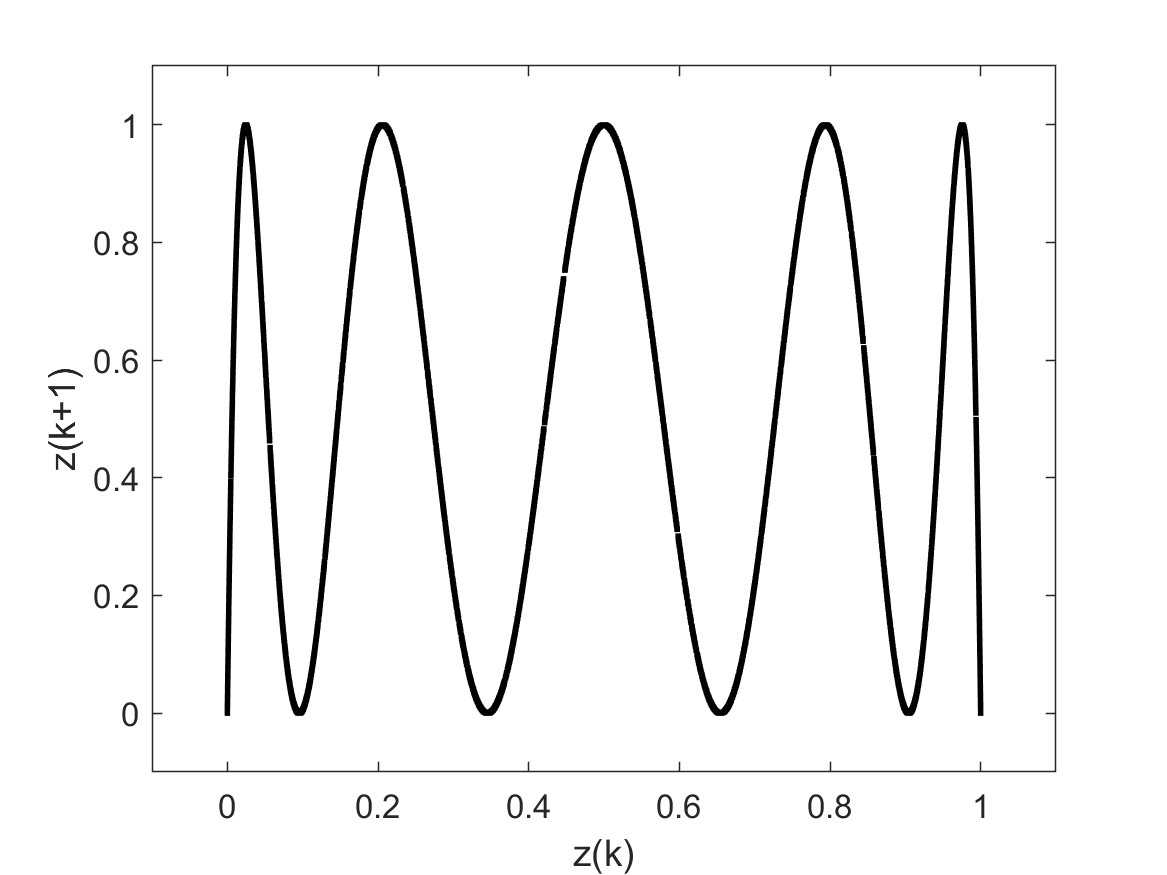}

(a) $\mathcal{B}(z,0.5,0.5)$\hspace{2.1cm} (d) Eq. \eqref{eq:schr}, $b=10$

\includegraphics[trim = 9mm 3mm 10mm 8mm,clip, width=6.2cm, height=4.5cm]{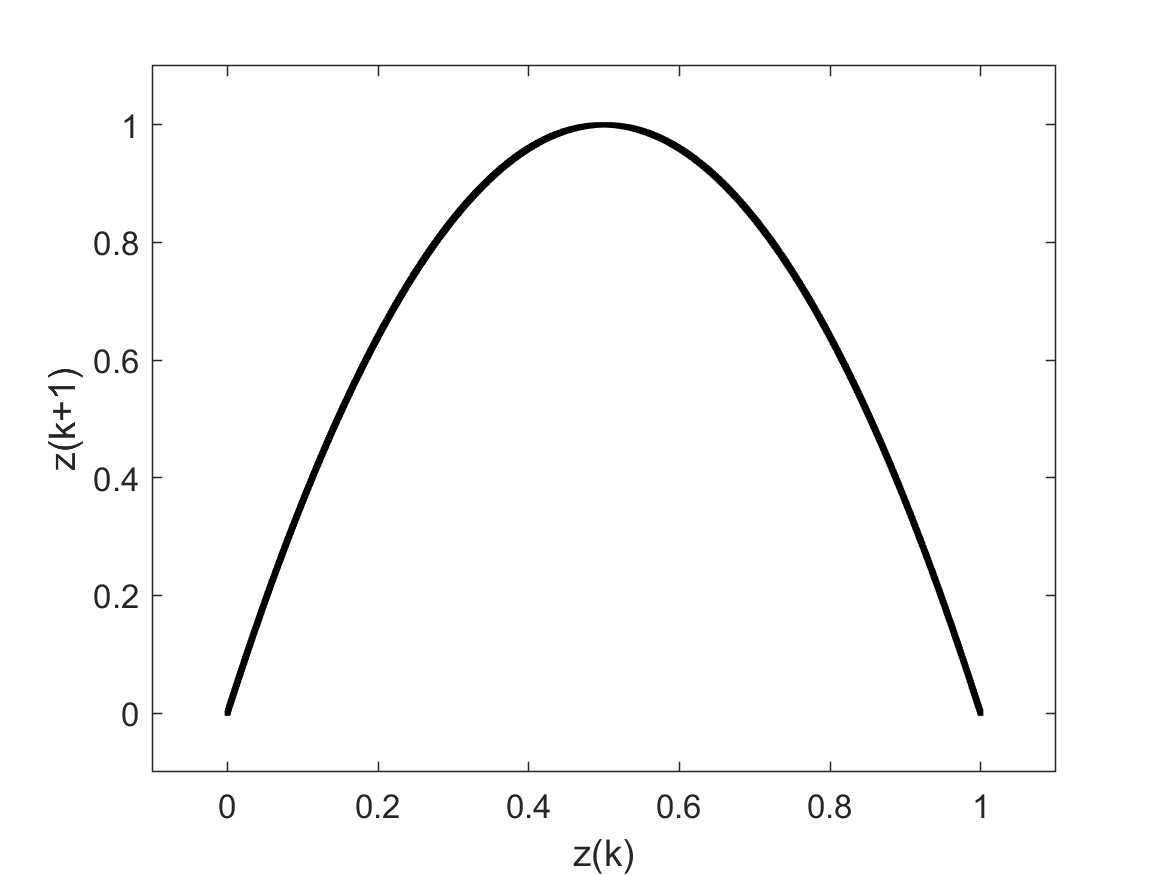}
\includegraphics[trim = 9mm 3mm 10mm 8mm,clip, width=6.2cm, height=4.5cm]{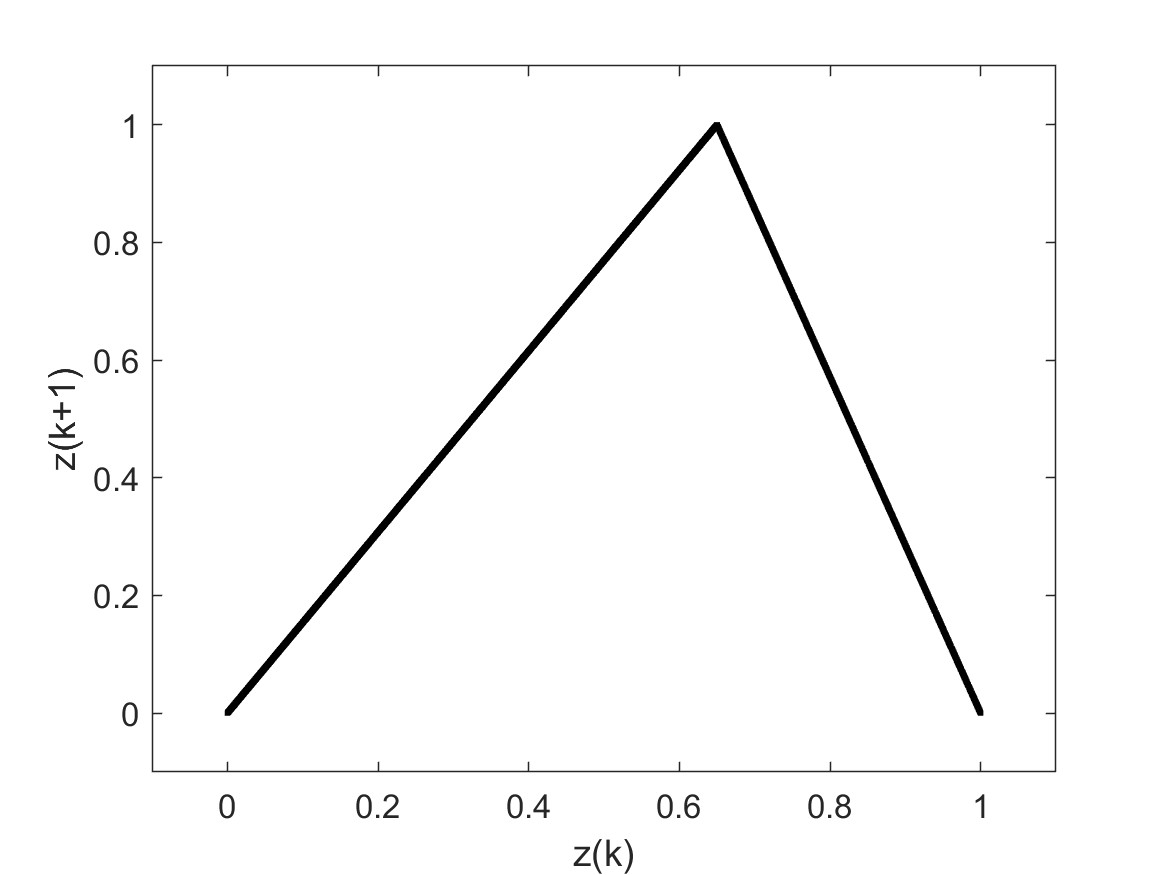}

 (b) Eq. \eqref{eq:schr}, $b=2$ \hspace{2cm} (e) Eq. \eqref{eq:atent}, $a=0.65$

 \includegraphics[trim = 9mm 3mm 10mm 8mm,clip, width=6.2cm, height=4.5cm]{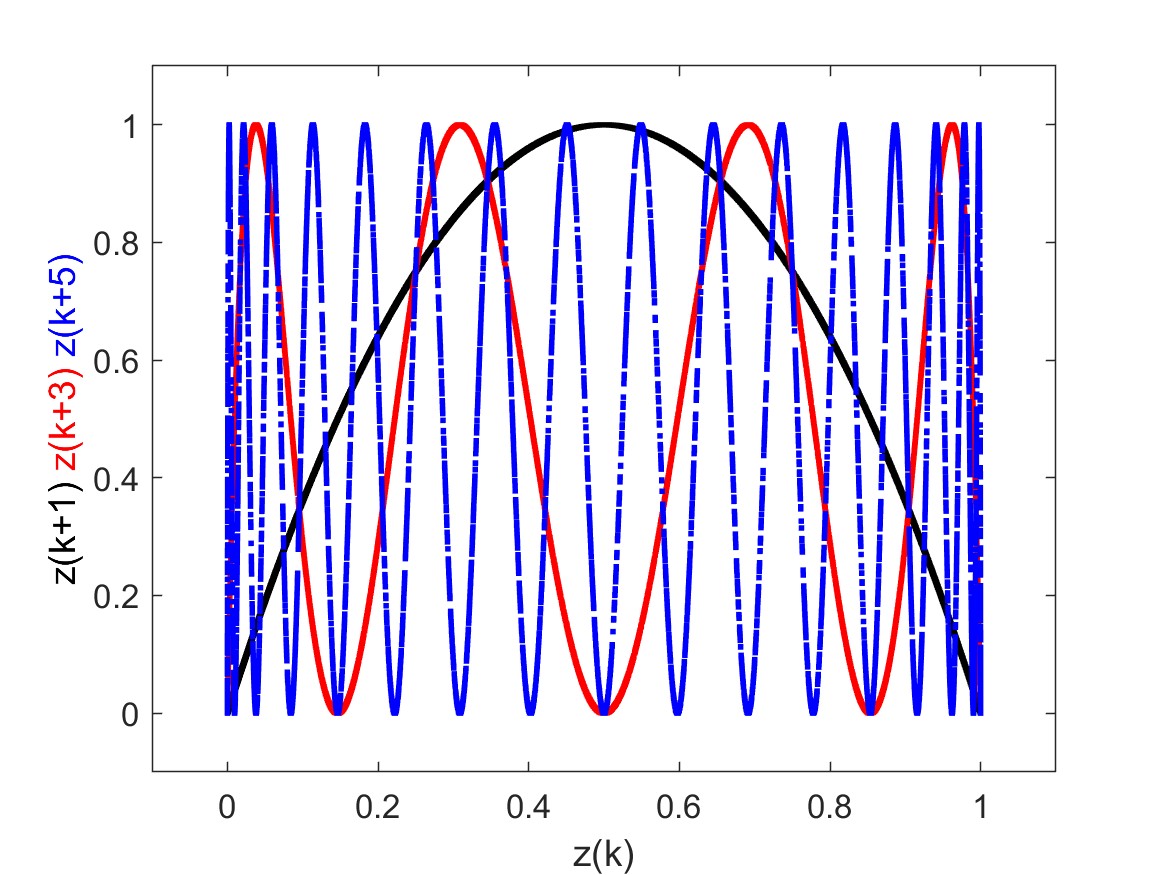}
\includegraphics[trim = 9mm 3mm 10mm 8mm,clip, width=6.2cm, height=4.5cm]{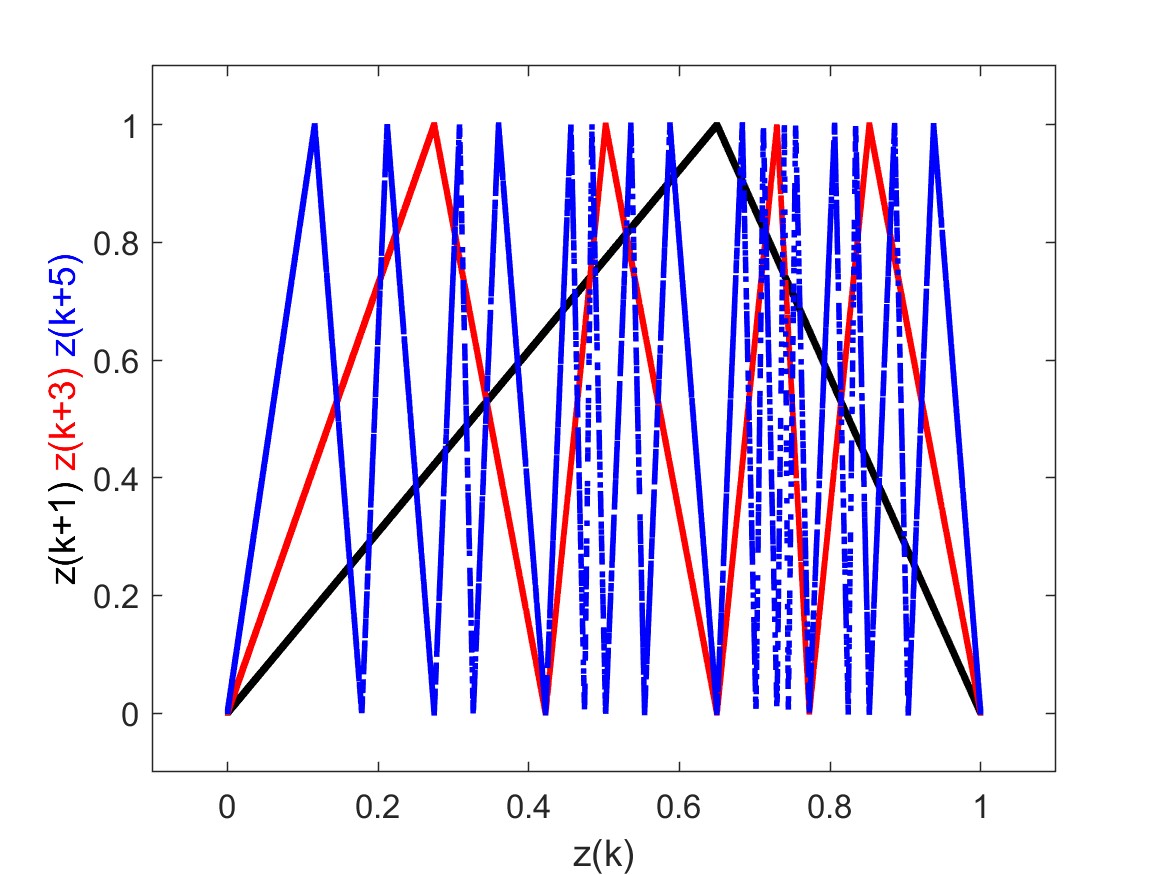}

 (c) Eq. \eqref{eq:schr}, $b=2$ \hspace{2cm} (f) Eq. \eqref{eq:atent}, $a=0.65$

\caption{Return maps of random and chaotic values. (a) Random values. (b), (d), (e) First-order return maps of chaotic values. (c), (f) Higher-order return maps of chaotic values.}
\label{fig_return}

\end{figure}

As mentioned earlier, a fundamental difference between random and chaotic values (and thus a potential source of apparent performance differences arising from their use) is the relationship between successive values. A convenient and illustrative way of visualizing the relationship of successive values is the return map $z(k+1)$ vs. $z(k)$, see e.g.~\cite{ott93}. Fig. \ref{fig_return} shows the return maps of random values, but also of chaotic values obtained by the Schröder map, Eq. \eqref{eq:schr} with $b=2$ (and thus by the identical Logistic map, Eq.  \eqref{eq:logist}, with $r=4$) and $b=10$ and the asymmetric tent map, Eq. \eqref{eq:atent}, with $a=0.65$. 

We see that for random, Fig.~\ref{fig_return}(a), a given value $z(k) \in [0,1]$ can be followed by $z(k+1)$ spread over the entire interval $[0,1]$. The values $z(k)$ and $z(k+1)$ are a slightly denser towards the boundaries of $[0,1]$, which is a consequence of the $\mathcal{B}(z,0.5,0.5)$ random depicted, which has an arcsine distribution, see Fig. \ref{fig_dist}(a).  For the return maps of chaotic values, we obtain a different picture, see Fig.~\ref{fig_return}(b),(d),(e). Here, the return map $z(k)$ vs. $z(k+1)$ has the form of the graph that belongs to the nonlinear map. Note that for the Schröder map  \eqref{eq:schr} with  $b=10$ the graph covers a greater proportion of the square $[0,1]^2$, compared to $b=2$. 

The return maps in Fig.~\ref{fig_return}(b),(d),(e) raise the question of how chaotic ergodicity and mixing may lead to a random-like coverage of $[0,1]^2$ comparable to that produced by random values, as illustrated in Fig.~\ref{fig_return}(a). This question can be addressed by examining higher-order return maps, see Fig. Fig.~\ref{fig_return}(c),(f) for the Schröder map  \eqref{eq:schr} with  $b=2$ and  the asymmetric tent map \eqref{eq:atent} with $a=0.65$, respectively. In addition to $z(k)$ vs. $z(k+1)$ as black dots, the higher-order return maps $z(k)$ vs. $z(k+3)$ and $z(k)$ vs. $z(k+5)$ are depicted as red and blue dots, respectively. We see that the square $[0,1]^2$ gradually fills as we take into account the dependencies of  values separated by more than one time step. These characteristics have consequences for an evolutionary search using either random or chaotic values. While random values direct the next search step approximately equally in all directions, chaotic values concentrate the search along directions dictated by the underlying nonlinear map. If such a directionality coincides with the search space structure, this property may lead to advantages in performance.    

In a sense, the quantitative measures of chaos discussed next provide a means of conceptualizing and quantifying the properties just
observed in the return map as they relate successive values to each other. The most obvious quantitative measure of chaos is the Lyapunov exponent (LE), which quantifies how two time series starting infinitesimally close to each other diverge, see e.g.~\cite{ott93}.
Unfortunately, the LE 
cannot be calculated for random time series (by definition, the LE is undefined (infinite) for (ideal) random sequences). As we also want to compare with random time series produced by PRNGs, we additionally consider two measures that can be calculated for both random and chaos, the mutual information (MI) and the permutation entropy (PE). 

For MI, we consider the time-delayed mutual information $\text{MI}(\tau)=\text{MI}(z(k),z(k+\tau))$ where $\tau \ge 1$ is an integer time lag. The quantity $\text{MI}(\tau)$ measures the dependency between the sequence and its delayed version at a specific lag $\tau$~\cite{fras86,kan97}:
\begin{equation}
\text{MI}(\tau)
=
\sum_{i=1}^{K}\sum_{j=1}^{K}
p_{ij}(\tau)
\log
\left(
\frac{p_{ij}(\tau)}
{p_i(\tau)\,p_j(\tau)}
\right) \label{eq:mi}
\end{equation}
where 
$p_{ij}(\tau)
=
\Pr\!\left\{z(k)=a_i,\;z(k+\tau)=a_j\right\}$ is the jointed probability distribution over an equidistant partition with discrete values $(a_1,a_2,\ldots,a_K)$ within the interval $[0,1]$ and
$p_i(\tau)
=
\sum_{j=1}^{K}p_{ij}(\tau)$ and
$
p_j(\tau)
=
\sum_{i=1}^{K}p_{ij}(\tau)$ are the marginal probabilities. The numerical results reported in Sec.~\ref{sec:results} use $\tau=2$ (and $K=32$).

The quantity PE measures the complexity and temporal ordering of a sequence by evaluating ordinal pattern diversity. 
For calculating PE, we put an embedding form vector
\begin{equation*} \mathbf{v}_i = \bigl(z(i), z(i+\tau), \dots, z(i+(d-1)\tau)\bigr), 
\quad i = 1, \dots, N - (d-1)\tau \end{equation*}
with pattern length (embedding dimension) $d \geq 2$ and lag $\tau \geq 1$. We define $S_d$ as the set of all permutations of $\{1,2,…,d\}$, and
 compute, for each $\mathbf{v}_i$, ordinal patterns by the permutation $\pi_i \in S_d$ that sort its components in ascending order with ties handled by a fixed rule. This produces a sequence of symbols of "words” from $d\mkern2mu !$ possible orderings:
\begin{equation*}z(i + (\pi_i(1)-1)\tau) \le z(i + (\pi_i(2)-1)\tau) \le \dots \le z(i + (\pi_i(d)-1)\tau),
 \end{equation*}
with $\pi_i \in S_d$.  From that we obtain an
empirical distribution with estimated probabilities $p_\pi(\tau,d)$ approximated by the relative frequencies of each pattern $\pi_i$ in the sequence:
\begin{equation}p_\pi(\tau,d) = \frac{1}{M} \sum_{i=1}^{M} \mathbf{1}\{\pi_i = \pi\}, 
\quad \pi \in S_d,
 \end{equation}
with $M = N - (d-1)\tau$, and finally get the
permutation entropy~\cite{ban02}:
\begin{equation} \mathrm{PE}(\tau,d) = -\sum_{\pi \in S_d} p_\pi(\tau,d)\,\log_2 p_\pi(\tau,d). \label{eq:pe} \end{equation} For numerical calculation of PE, we use $\tau=1$ and $d=5$ (and $N=10\:000$). For a discussion about the relationships and correlations between MI, PE and $\tau$, also for varying $\tau$, see the supplemental material.

\section{Experimental Setup} \label{sec:setup}

\subsection{PSO Algorithm} \label{sec:pso}
Particle swarm optimization (PSO) searches for an optimum by iteratively updating a population of candidate solutions.   The swarm consists of $I$ particles operating in a $D$-dimensional search space. For each particle $i=1,2,\ldots, I$ at iteration $t$, the next position $x_i(t+1)$ is determined by its current position $x_i(t)\in\mathbb{R}^D$, its velocity $v_i(t)\in\mathbb{R}^D$, the (local) best position it has encountered up to that point, denoted by $lp^t_i$, and the (global) best position discovered by the entire swarm up to iteration $t$, denoted by $gp^t$. In addition, the next position $x_i(t+1)$  depends directly on random-like components $r_1$ and $r_2$, which  are obtained from sequences $z(k)$ generated by either PRNGs or chaotic maps. In this study, a standard inertia-weight PSO is used, which updates
the position and velocity according to
\begin{align}
x_i(t+1) &= x_i(t) + v_i(t+1), \label{eq:PSO1} \\
v_i(t+1) &= wv_i(t) + c_1r_1\bigl(lp^t_i-x_i(t)\bigr)
+ c_2r_2\bigl(gp^t-x_i(t)\bigr). \label{eq:PSO2}
\end{align}
The weight $w$ controls the contribution of the particle's previous velocity, while $c_1$ and $c_2$ determine the strength of the cognitive and social contributions, respectively. For every run the initial population and velocity is generated uniformly at random in the intervals $[-30,30]$ and $[-6,6]$, respectively. These bounds are also observed by position bounds and velocity clamping during the run.    
 We take standard values of PSO parameters frequently used:
 $w = 0.79$, $c_1 = c_2 = 1.49$ and $I=60$  particles per swarm.

 \subsection{Random and Chaotic Sequences}
In the numerical experiments, we compare random and chaotic sequences $z(k)$. The random values according to the Beta distributions $\mathcal{B}(z,0.5,0.5)$ (arcsine) and  $\mathcal{B}(z,1,1)$ (uniform) are obtained by a Mersenne Twister PRNG. The chaotic values are obtained from the Schröder map, Eq. \eqref{eq:schr}, with the parameter $b=\{2,3,4,6,8,10\}$ (arcsine), and from the asymmetric tent map, Eq. \eqref{eq:atent}, with the parameter $a=\{0.5,0.55,0.6,0.65,0.7,0.8,0.9\}$ (uniform).   
Due to the ergodicity of chaotic behavior, 
the maps can be seeded by different $x(0) \in [0,1]$.
Each particle for each of its D dimensions uses its own sequence from the one-dimensional chaotic maps seeded for each run by a uniformly random $x(0)$. This is done separately for $r_1$ and $r_2$.  Calculating LE, MI and PE, see Sec.~\ref{sec:temp_diff}, uses the same sequences.

\subsection{Benchmark Problems} \label{sec:BBOB}
We consider the 24 functions in the BBOB benchmark suite~\cite{hans10}. To account for the variability introduced by the problem transformations used in BBOB, each function is evaluated on 3 distinct instances. A BBOB instance represents a particular realization of a benchmark function obtained through randomized transformations that preserve its characteristic properties. For every instance, we conduct 15 statistically independent PSO runs, using separate seeds for the random or chaotic sequences employed by the algorithm. Consequently, each benchmark function is assessed over 45 runs in total. The numerical budget is fixed at $10\:000$ function evaluations for every individual run. We consider each benchmark function assessment for three search space dimensions: $D=\{10,30,50 \}$.

\subsection{Performance Metrics}
For evaluating the differences in PSO performance over random and chaotic values with arcsine and uniform distribution, we use an anytime performance assessment~\cite{hans22,ye22,lo24}. For the results obtained with the BBOB benchmark suite (see Sec.~\ref{sec:BBOB}), we employ the so--called empirical cumulative distribution function (ECDF) which records the number of targets attained at a given numerical budget. In numerical experiments we approximate the ECDF by observing performance data to obtain an empirical attainment function (EAF). Furthermore, we use the area under the EAF curve (AUC) as a quantification of the PSO anytime performance. Following common practice~\cite{lo24,niki24}, all experiments use a target interval of $[10^{-8},10^2]$ and a numerical budget of $10\:000$ function evaluations.

 \section{Results and Discussion} \label{sec:results}

 \subsection{Anytime Performance by EAF Curves} \label{sec:res_any}

\begin{figure}

\includegraphics[trim = 9mm 3mm 10mm 8mm,clip, width=6.2cm, height=4.5cm]{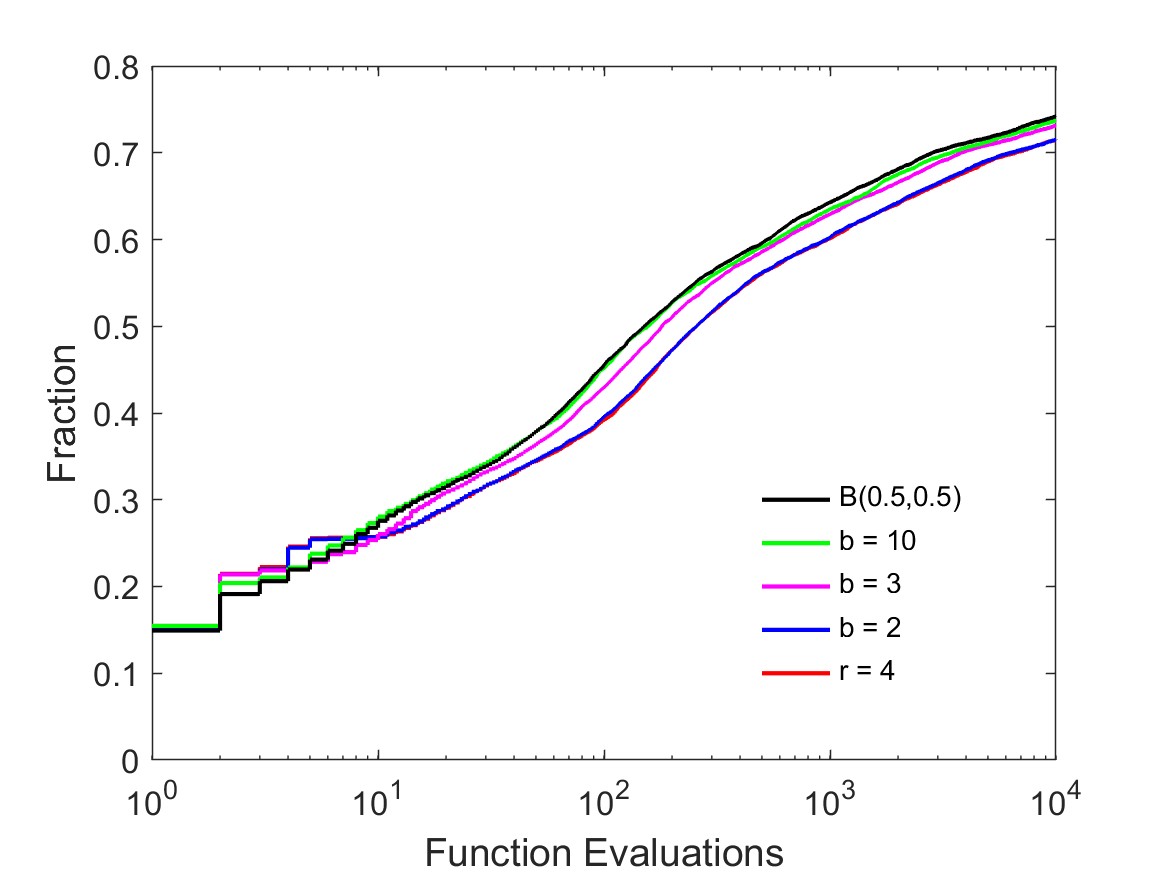}
\includegraphics[trim = 9mm 3mm 10mm 8mm,clip, width=6.2cm, height=4.5cm]{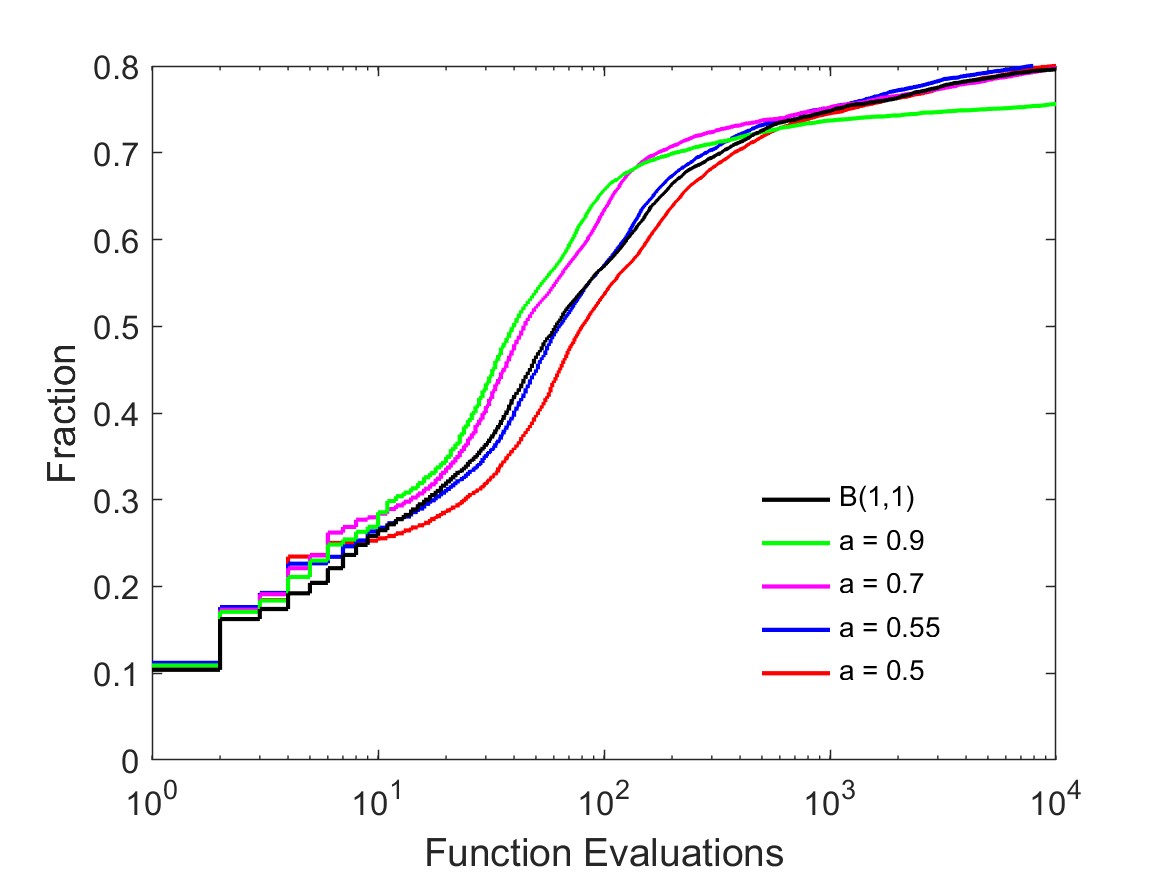}

$D=10$: \hspace{0.3cm} (a) Arcsine  \hspace{2.2cm} (d) Uniform \hspace{1.3cm} 

\includegraphics[trim = 9mm 3mm 10mm 8mm,clip, width=6.2cm, height=4.5cm]{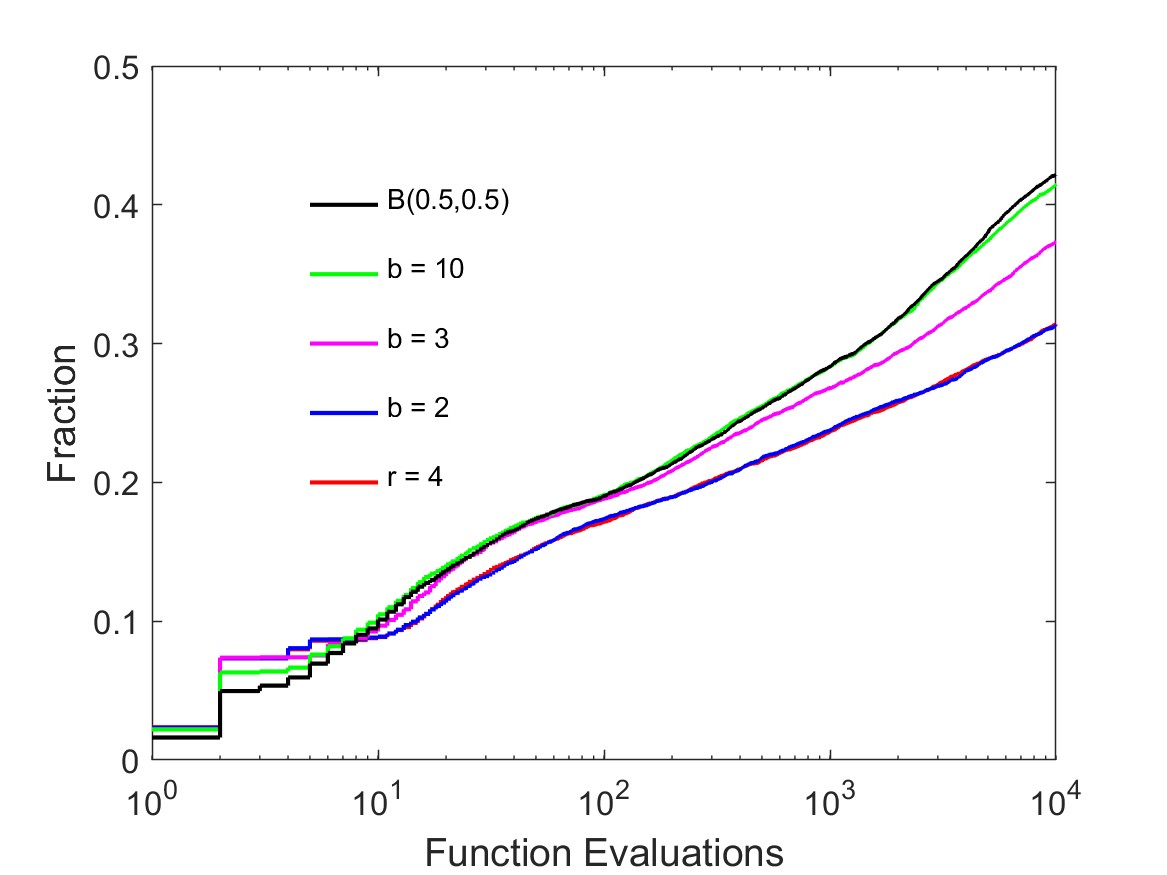}
\includegraphics[trim = 9mm 3mm 10mm 8mm,clip, width=6.2cm, height=4.5cm]{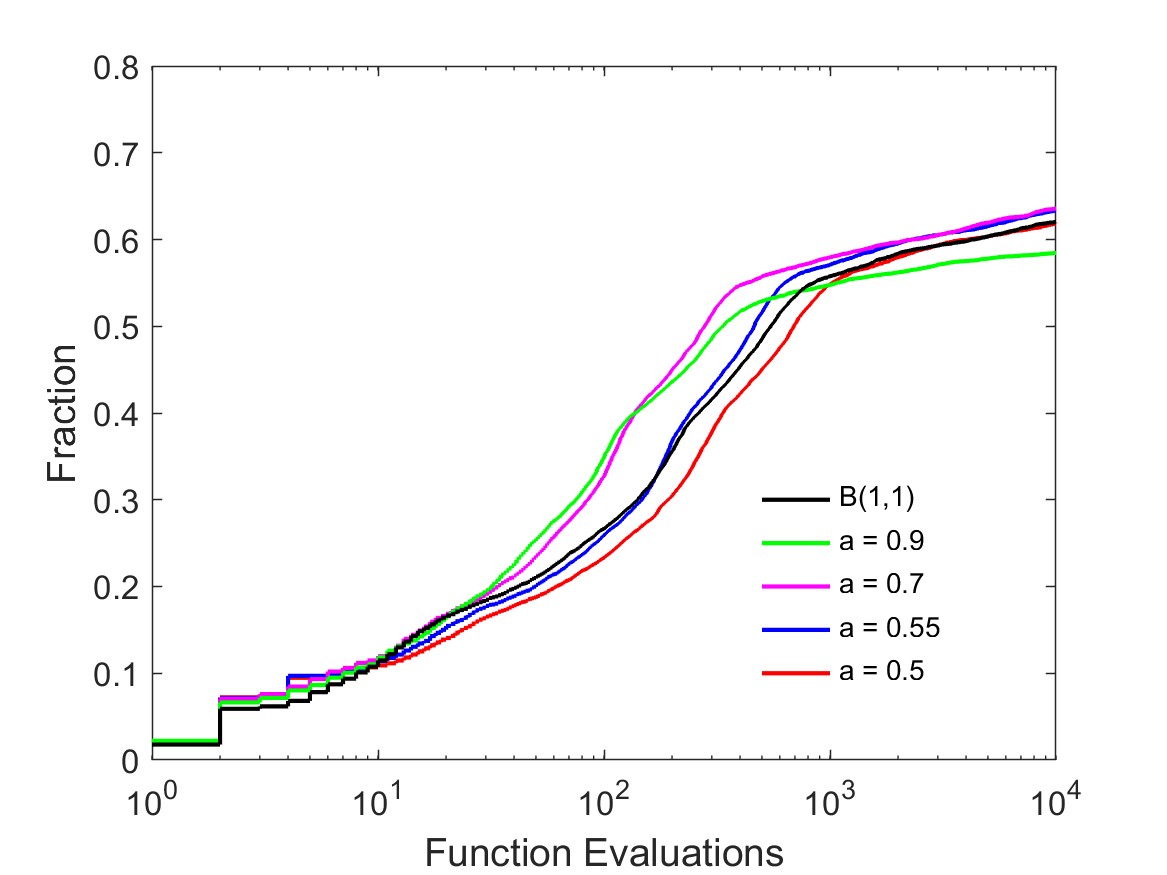}

$D=30$: \hspace{0.3cm} (b) Arcsine  \hspace{2.2cm} (e) Uniform \hspace{1.3cm}

\includegraphics[trim = 9mm 3mm 10mm 8mm,clip, width=6.2cm, height=4.5cm]{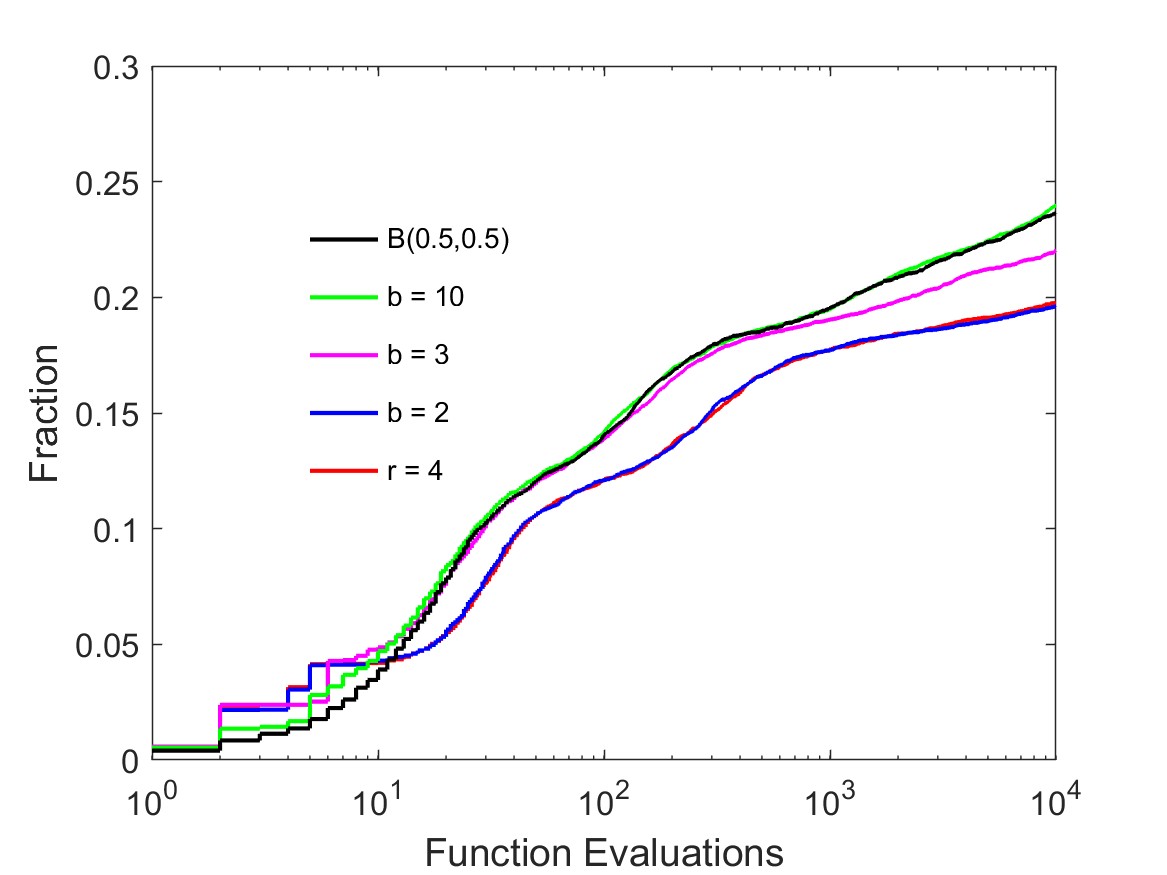}
\includegraphics[trim = 9mm 3mm 10mm 8mm,clip, width=6.2cm, height=4.5cm]{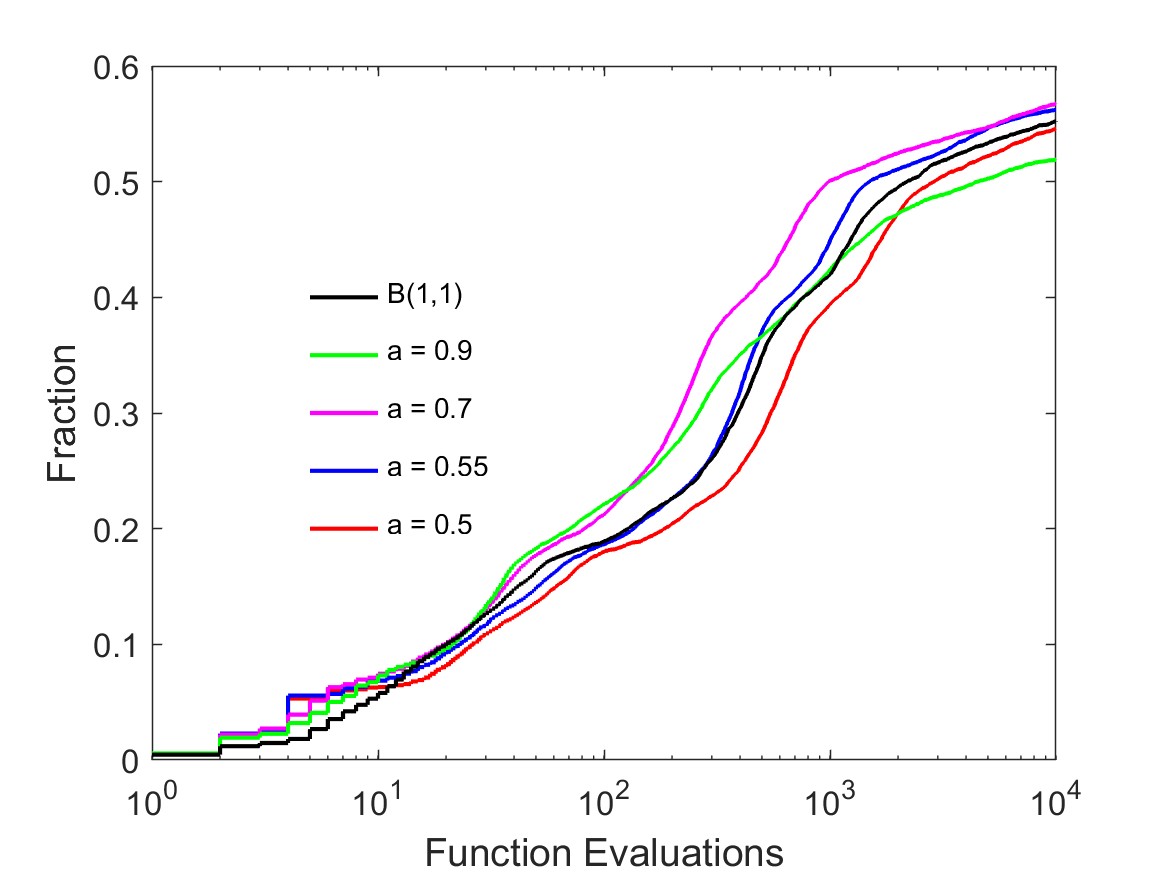}

$D=50$: \hspace{0.3cm} (c) Arcsine  \hspace{2.2cm} (f) Uniform \hspace{1.3cm} 
 
\caption{Results for chaotic vs. random sequences by EAFs quantifying the proportion of runs in which an arbitrary target value is reached within a specified numerical budget.  Chaotic values for the arcsine distribution are obtained by the Schröder map \eqref{eq:schr}, while for the uniform distribution we take the asymmetric tent map \eqref{eq:atent}. Search space dimension $D=\{10,30,50 \}$. The results are averages over the complete BBOB benchmark suite and 15 independent runs for 3 instances each.}
\label{fig_anytime}

\end{figure}

\begin{figure}

\includegraphics[trim = 5mm 3mm 10mm 8mm,clip, width=6.2cm, height=4.5cm]{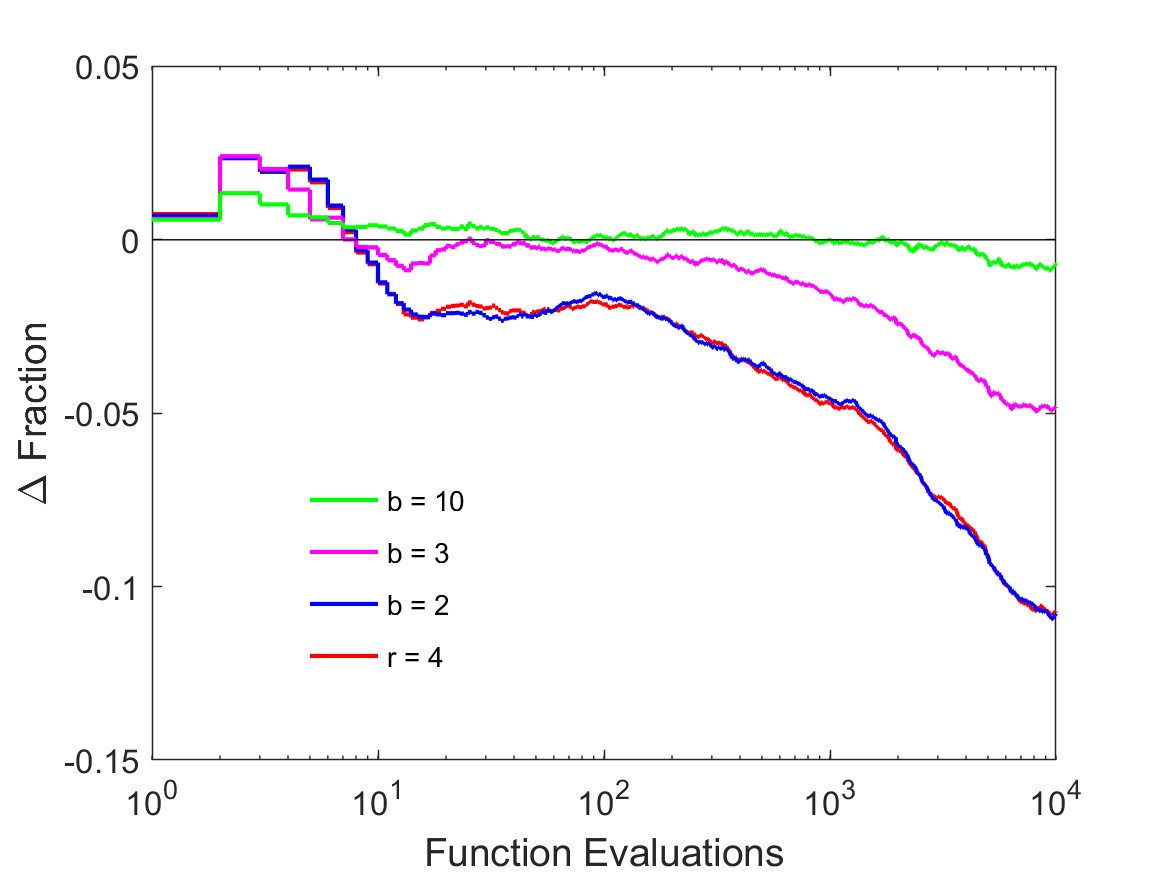}
\includegraphics[trim = 5mm 3mm 10mm 8mm,clip, width=6.2cm, height=4.5cm]{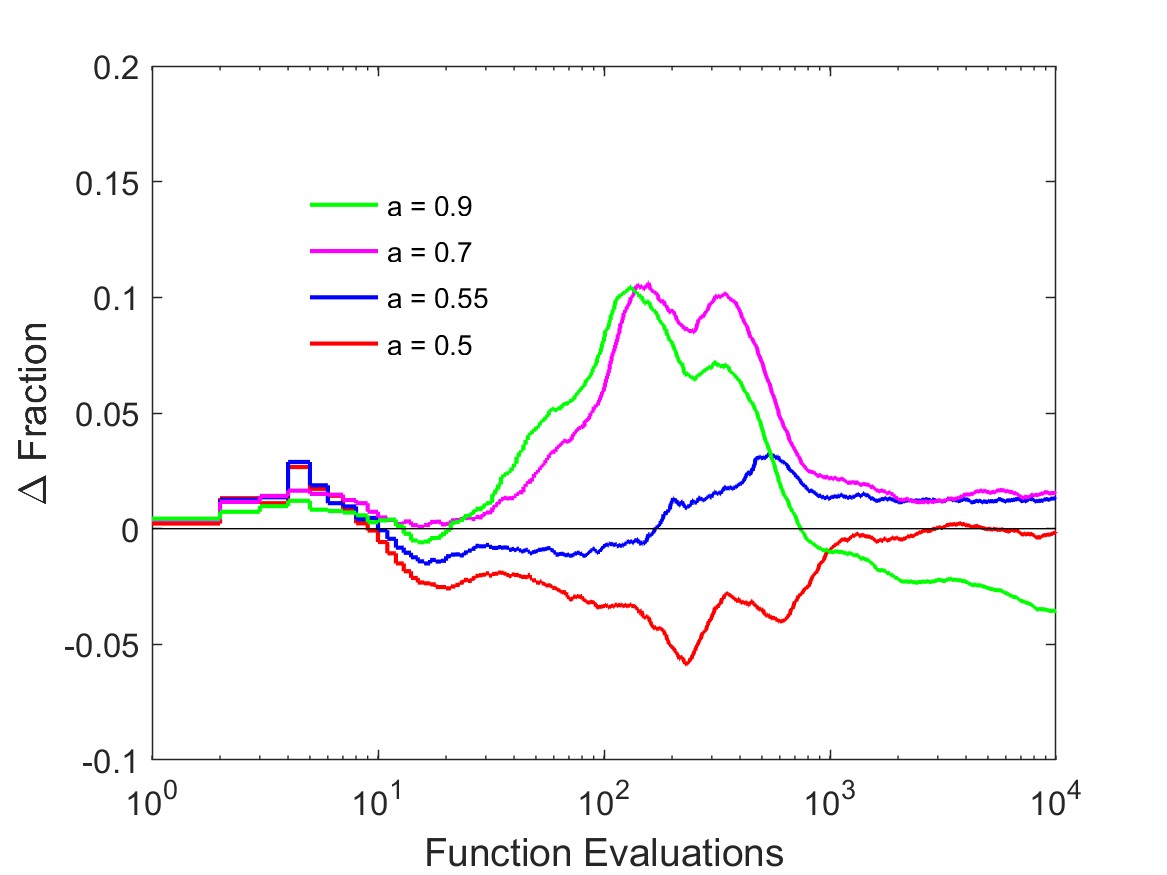}

BBOB 1-24: \hspace{0.3cm} (a) Arcsine  \hspace{2.2cm} (d) Uniform \hspace{1.3cm} 

\includegraphics[trim = 5mm 3mm 10mm 8mm,clip, width=6.2cm, height=4.5cm]{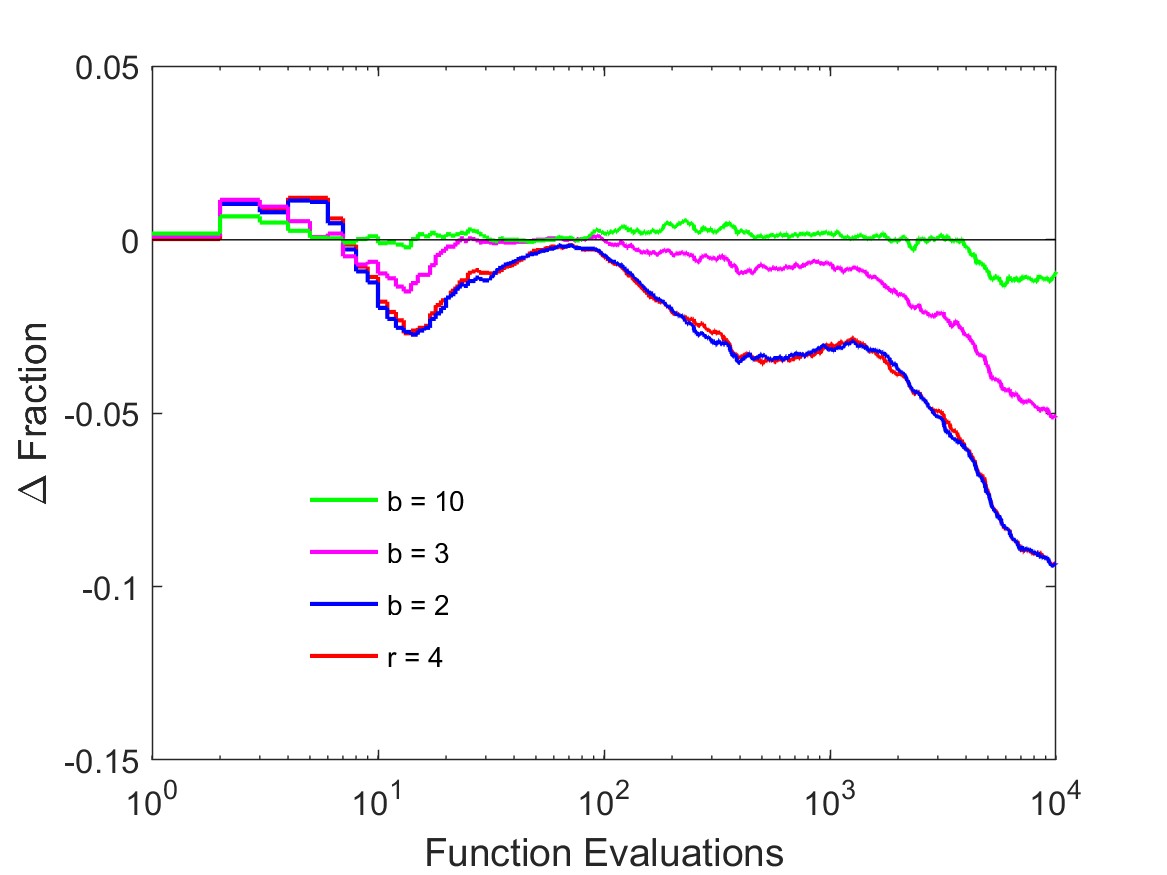}
\includegraphics[trim = 5mm 3mm 10mm 8mm,clip, width=6.2cm, height=4.5cm]{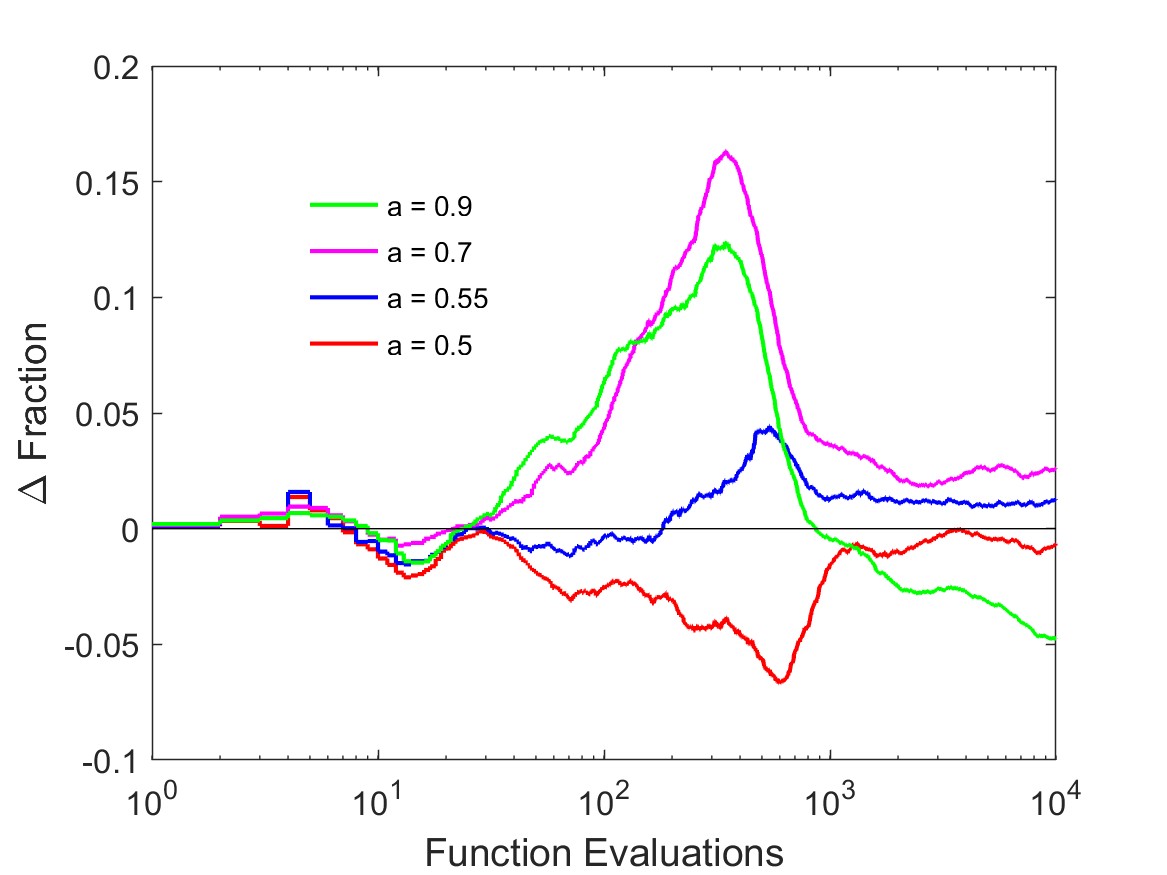}

BBOB 1-14: \hspace{0.3cm} (b) Arcsine  \hspace{2.2cm} (e) Uniform \hspace{1.3cm}

\includegraphics[trim =7mm 3mm 10mm 8mm,clip, width=6.2cm, height=4.5cm]{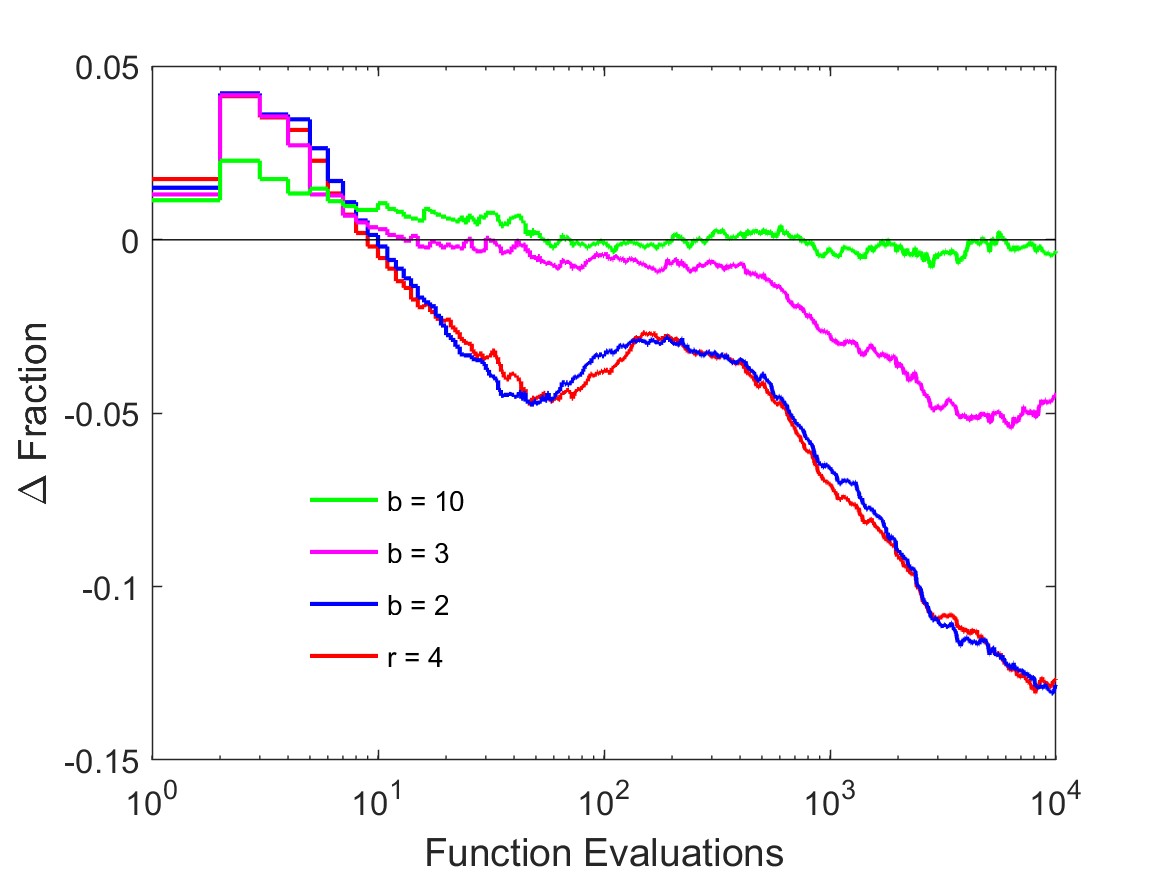}
\includegraphics[trim = 5mm 3mm 10mm 8mm,clip, width=6.2cm, height=4.5cm]{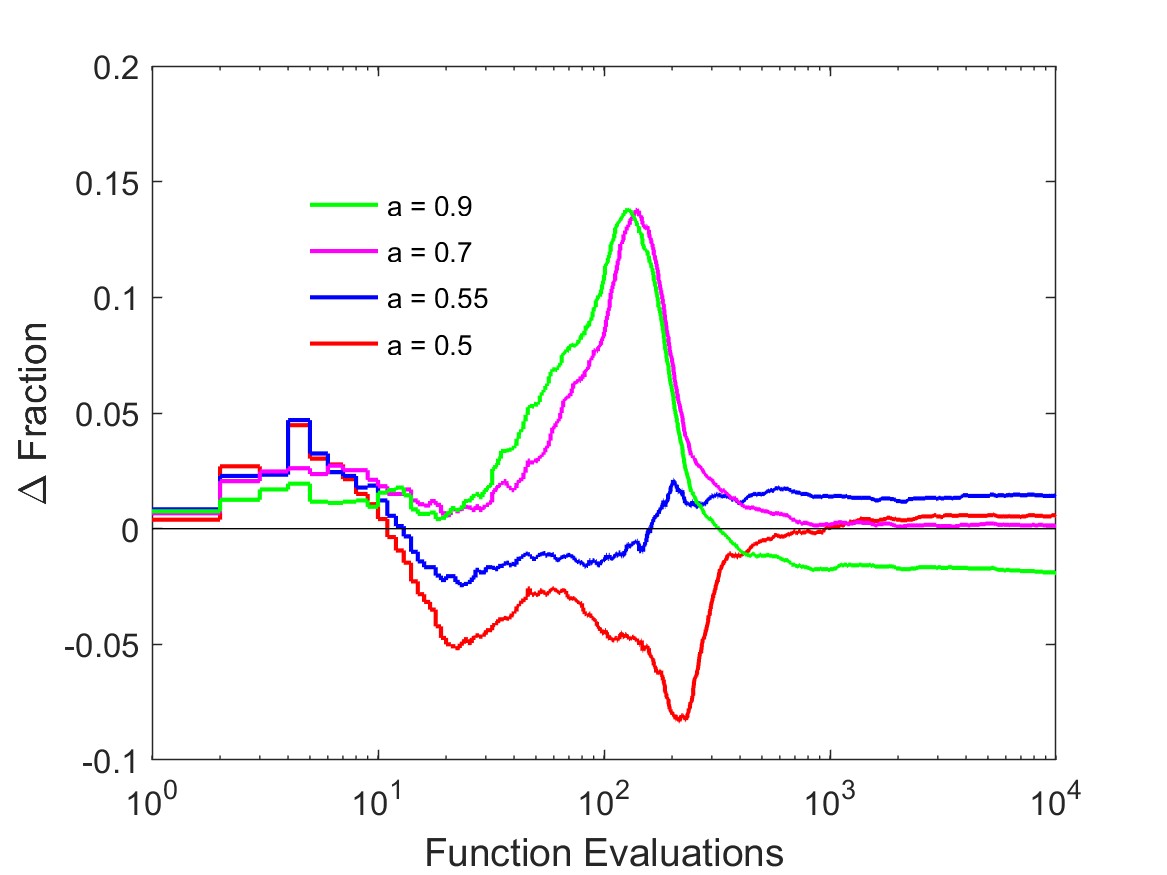}

BBOB 15-24: \hspace{0.2cm} (c) Arcsine  \hspace{2.2cm} (f) Uniform \hspace{1.3cm} 
 
\caption{Differences between the EAF curves for random and chaotic values by the quantity $\Delta \: \text{Fraction} = \text{EAF}_{chao}-\text{EAF}_{rand}$. Averages over the complete BBOB benchmark suite and 15 independent runs for 3 instances each. Search space dimension $D=30$. (a), (d) all BBOB functions, (b), (e) unimodal BBOB functions, (c), (f) multimodal BBOB functions. }
\label{fig_anytime_comp}

\end{figure}

Using the experimental setup  given in the previous section, we now present, analyze and discuss numerical results. Fig.~\ref{fig_anytime} gives anytime performance results by the EAFs. The $y$-axis shows the cumulative  fraction of runs reaching a given target value, while the $x$-axis indicates the corresponding runtime expressed by the number of function evaluations. From the EAF curves, we may deduce how well random and chaotic values contribute to performance for the PSO implementation detailed in Sec. \ref{sec:pso}. The curves are for the complete runtime of $10\:000$ function evaluations and give averages over the complete BBOB benchmark suite and 15 independent runs for 3 instances each. Thus, they average over $24 \times 15 \times 3=1080$ runs,  see the details in Sec. \ref{sec:BBOB}. Covering the search space dimensions $D=\{10,30,50\}$, Fig.~\ref{fig_anytime}(a)-(c) shows results for  arcsine-distributed values, while  Fig.~\ref{fig_anytime}(d)-(f) reports outcomes for uniform-distributed values. 

 A first observation is that uniformly distributed values consistently yield better results than arcsine-distributed values throughout the runtime. The differences are still somewhat subtle for $D=10$, but become substantially larger for $D=30$, and even more so for $D=50$, see also Sec.~\ref{sec:chao_auc} and Tab.~\ref{tab:boots} for quantification. We see that the performance drop for the search space dimension becoming larger (which is an expected outcome, particularly given that the swarm size $I$ remains fixed) is much more prominent in arcsine-distributed values than in uniform-distributed values. This result confirms previous findings that other chaotic maps, for instance the asymmetric tent map, often outperform the classic Logistic map (which produces arcsine values) on the same PSO variant~\cite{aro14,rong10}. If we look at the arcsine values, we see that we get almost the same result for $b=2$ and $r=4$. Given the identity of the Schröder map and the Logistic map, see the supplemental material,  for these parameter values, the results represent the level of difference obtained for runs repeated with the same parameter setting.   

\subsection{EAF Curve Differences between Random and Chaos}

 For a more detailed analysis of differences between random and chaos, Fig. 
\ref{fig_anytime_comp} gives the quantity
$\Delta \: \text{Fraction}$ for the search space dimension $D=30$ (results for $D=\{10,50\}$ can be found in the supplemental material). This quantity is the differences between the EAF curves for random and chaotic values: $\Delta \: \text{Fraction} = \text{EAF}_{chao}-\text{EAF}_{rand}$ for parameter values of the chaotic maps considered. We may interpret these curves as an indicator of how different the PSO performance is at certain runtime points over the whole numerical budget. Negative values of $\Delta \: \text{Fraction}$ indicate advantages for random values, while positive values suggest advantages for chaotic values. Fig. \ref{fig_anytime_comp}(a) and (d) give the  quantity $\Delta \: \text{Fraction}$ for all 24 functions of the BBOB benchmark suite. We see that  for all parameter settings of the chaotic maps, $\Delta \: \text{Fraction}$ changes its sign at least once over at least a short interval of the numerical budget. This strongly indicates that the question of whether random or chaotic values are preferable cannot be answered independently of the numerical budget.

Particularly interesting is that for both arcsine and uniform distributions chaotic values are good for small budgets, but the advantage frequently deteriorates for the budgets getting larger. This is most clearly the case for arcsine-distributed values, Fig. \ref{fig_anytime_comp}(a), where only $b=10$ remains in the vicinity  of random for the numerical budget advancing on $10\:000$ function evaluations.  A slightly different picture emerges for uniformly distributed values. At first, the differences are substantially larger and also for $a=0.7$ and $a=0.55$ the results remain better as the budget increases. For $a=0.9$, we have better results for a substantial budget interval, but for very large budgets, random overtakes. 

To study the effect that selecting benchmarks has on the anytime performance, we give the quantity $\Delta \: \text{Fraction}$ for subsets of the BBOB benchmark suite. Fig. \ref{fig_anytime_comp}(b) and (e) give the results for the BBOB functions 1-14, while Fig. \ref{fig_anytime_comp}(c) and (f) are for the BBOB functions 15-24. The selection is motivated by the supposed problem hardness of the benchmarks, given that the BBOB functions 1-14 are unimodal, while the functions 15-24 are multimodal. 

Mostly, the features observed for all 24 BBOB functions are also visible for the unimodal and multimodal subsets. For instance, chaos seems to be particularly advantageous for small budgets. This is even more the case for multimodal functions. Both subsets exhibit a deterioration in the performance of chaotic values, which is particularly marked for the arcsine distribution and multimodal functions. 
The results for uniformly distributed values present a more differentiated picture. For example, for multimodal functions and large budgets, chaos is better for all chaotic values except $a=0.9$, and the decrease in performance for all 24 BBOB functions is due to unimodal functions. Furthermore, the chaotic values for $a=0.7$ and $a=0.9$ produce peaks for intermediate budgets that level out for budgets becoming large. Moreover, these peaks in performance difference between random and chaos occur at slightly different intermediate budgets for unimodal vs. multimodal BBOB functions. Basically, these results are also obtained for $D=\{10,50\}$, see the supplemental material for the figures and some short additional discussion.  

\begin{figure}

\includegraphics[trim = 9mm 3mm 5mm 8mm,clip, width=6.2cm, height=4.5cm]{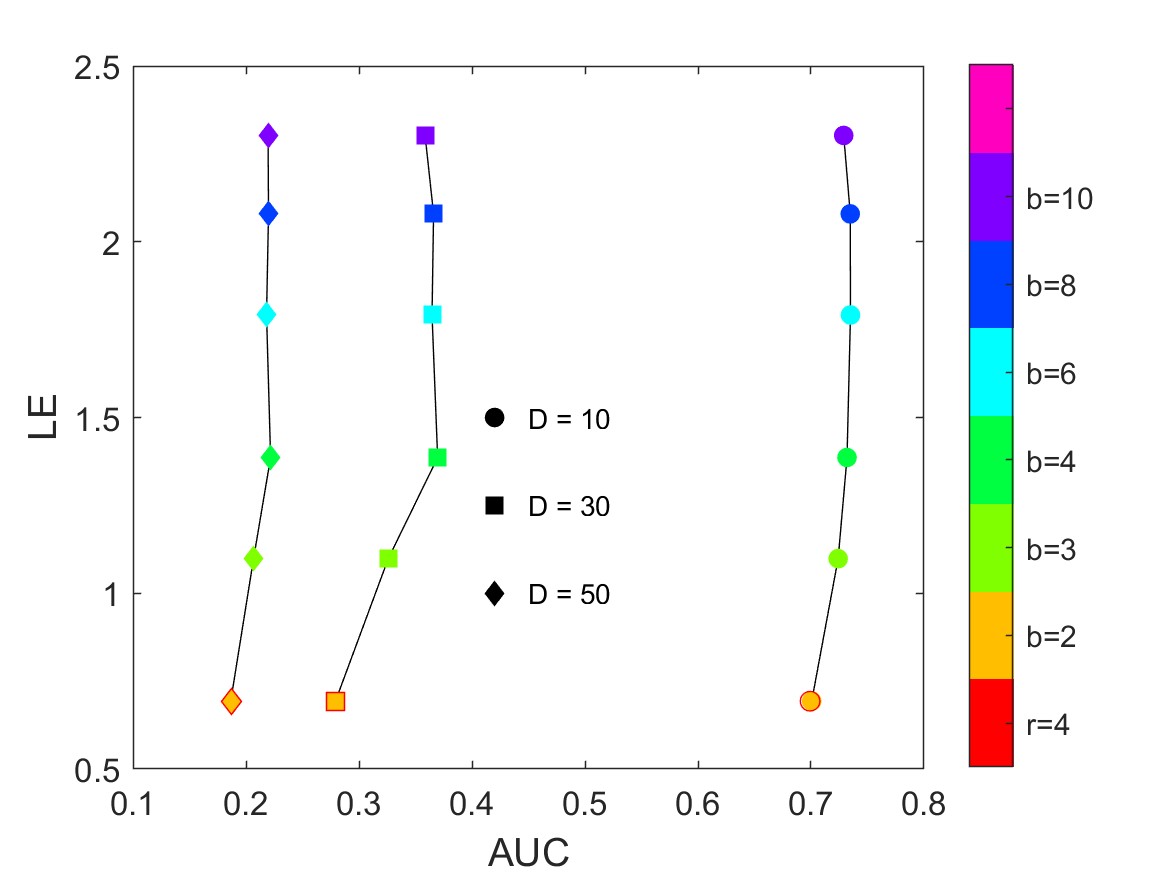}
\includegraphics[trim = 9mm 3mm 10mm 8mm,clip, width=6.2cm, height=4.5cm]{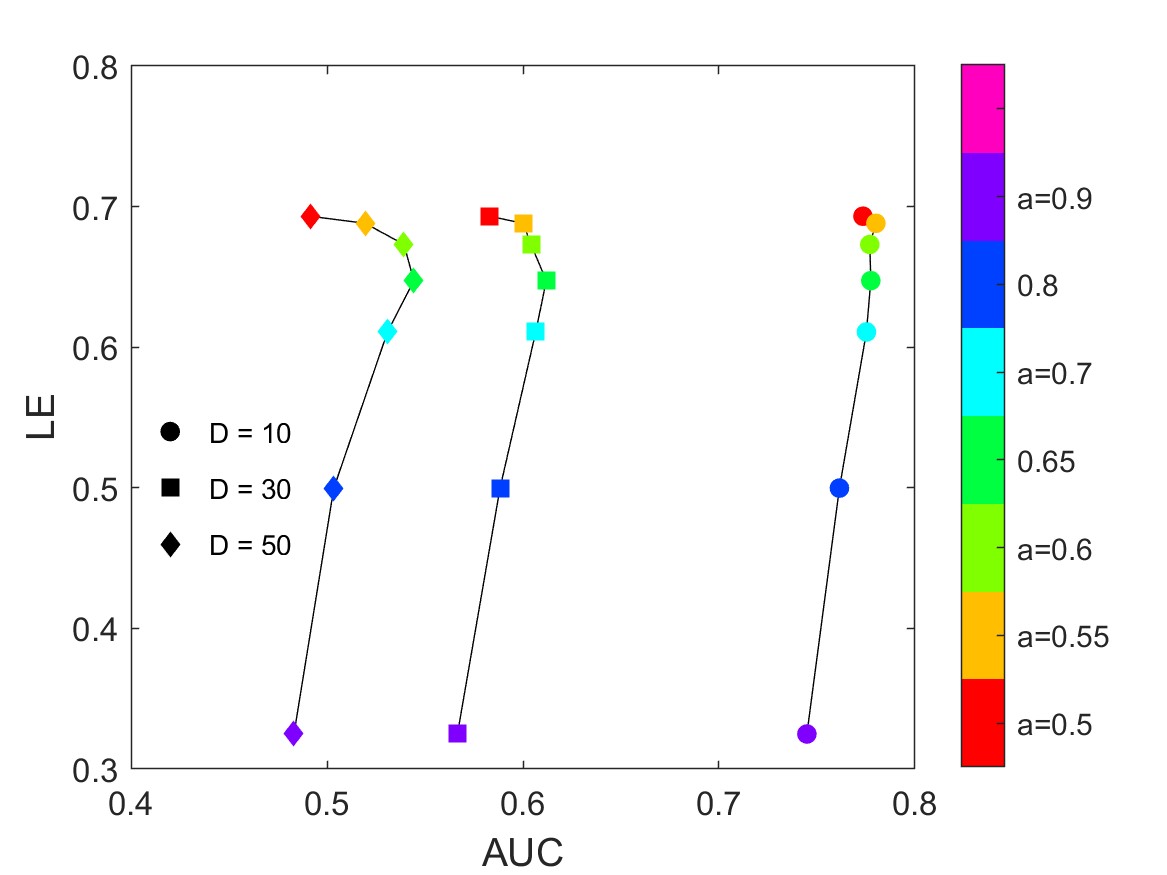}

 LE: \hspace{0.3cm} (a) Arcsine \hspace{3cm} (d) Uniform \hspace{1.2cm}

\includegraphics[trim = 9mm 3mm 5mm 8mm,clip, width=6.2cm, height=4.5cm]{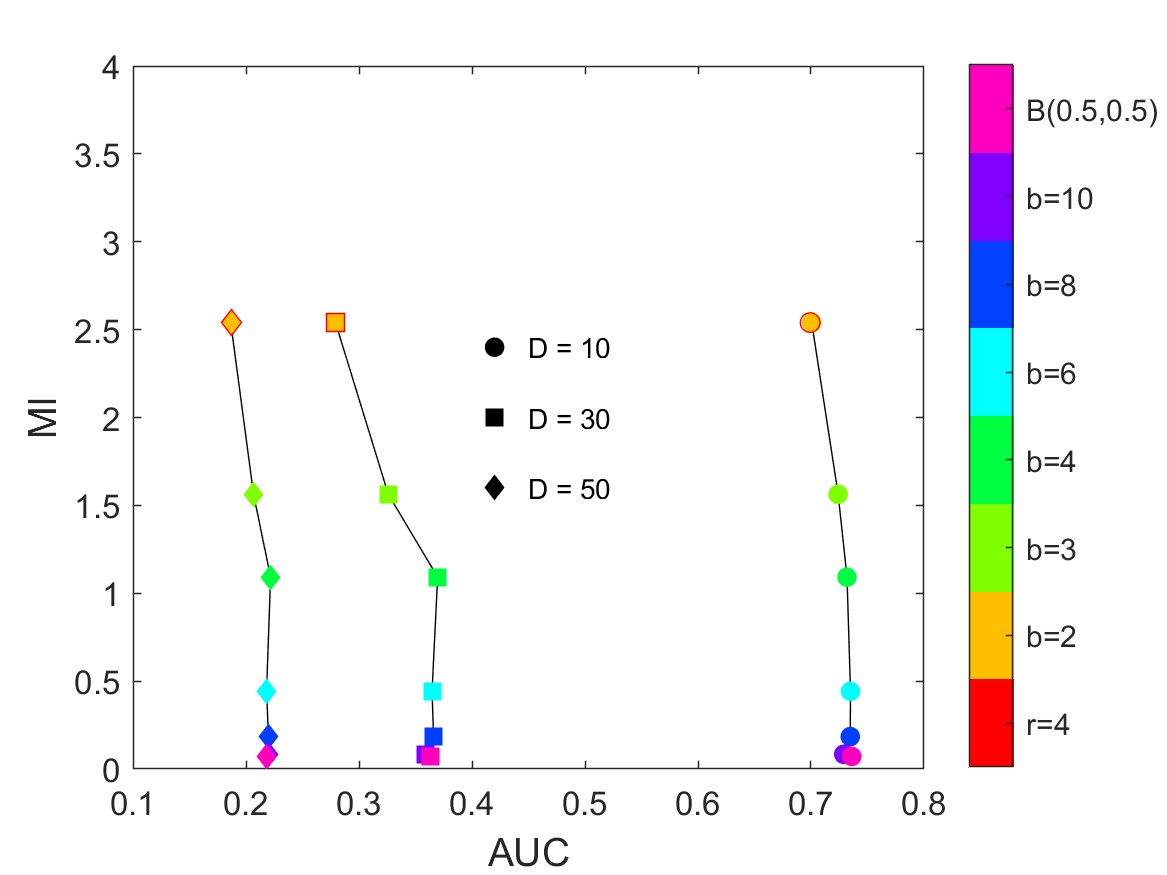}
\includegraphics[trim = 9mm 3mm 5mm 8mm,clip, width=6.2cm, height=4.5cm]{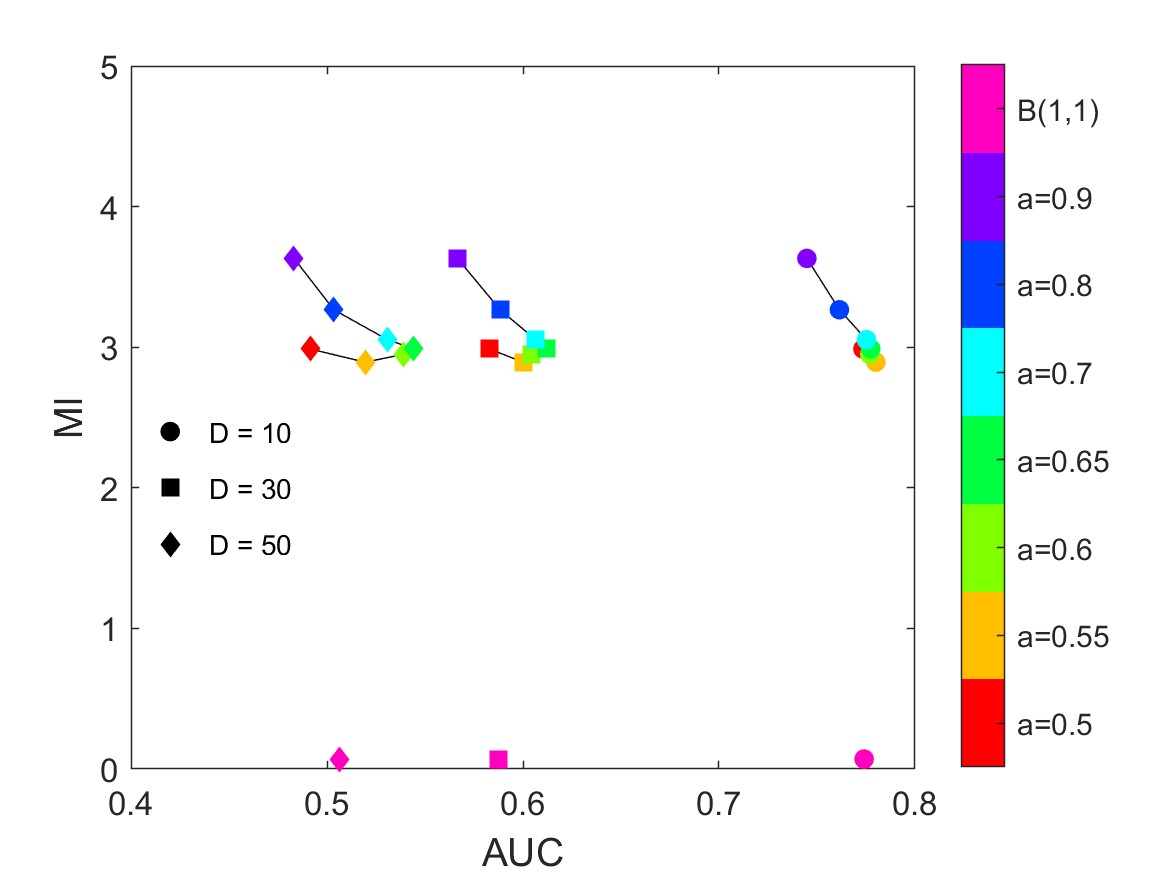}

MI: \hspace{0.3cm} (b) Arcsine \hspace{3cm} (e) Uniform \hspace{1.2cm}

\includegraphics[trim = 9mm 3mm 5mm 8mm,clip, width=6.2cm, height=4.5cm]{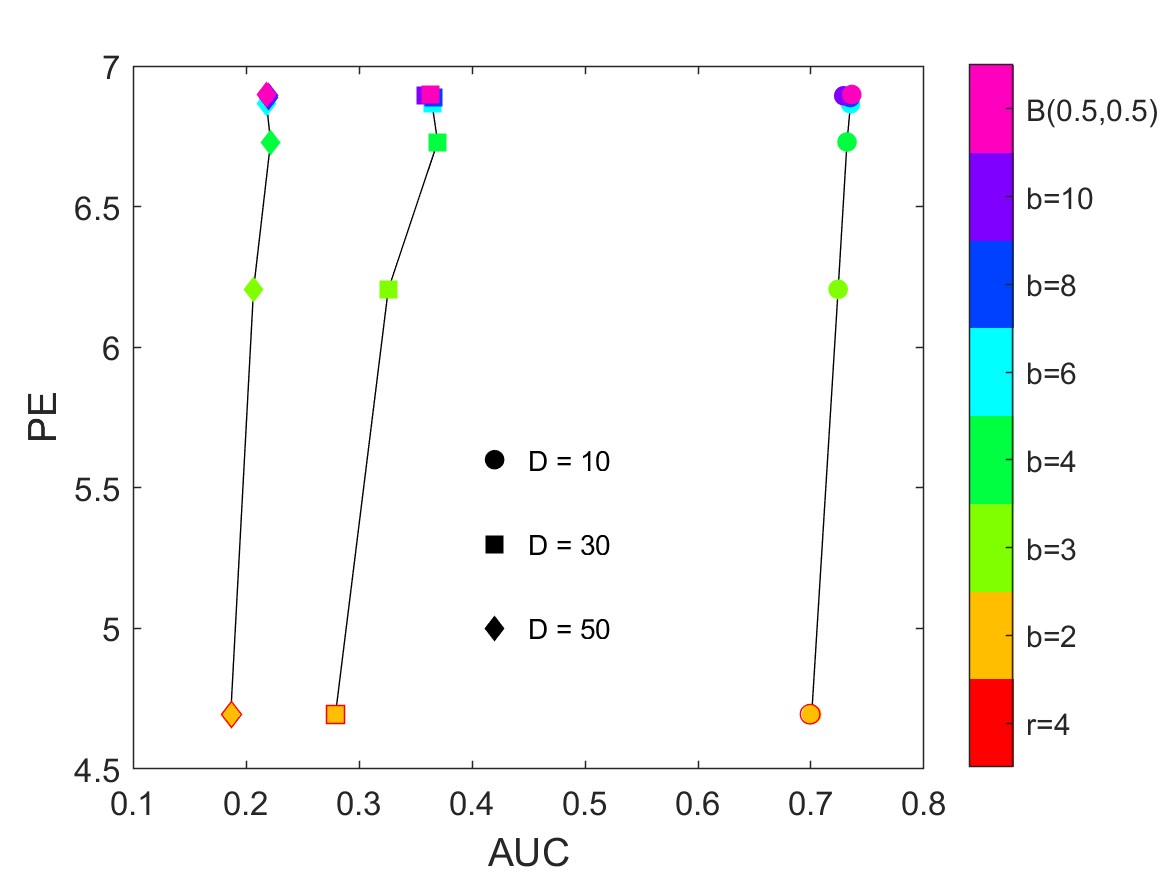}
\includegraphics[trim = 9mm 3mm 5mm 8mm,clip, width=6.2cm, height=4.5cm]{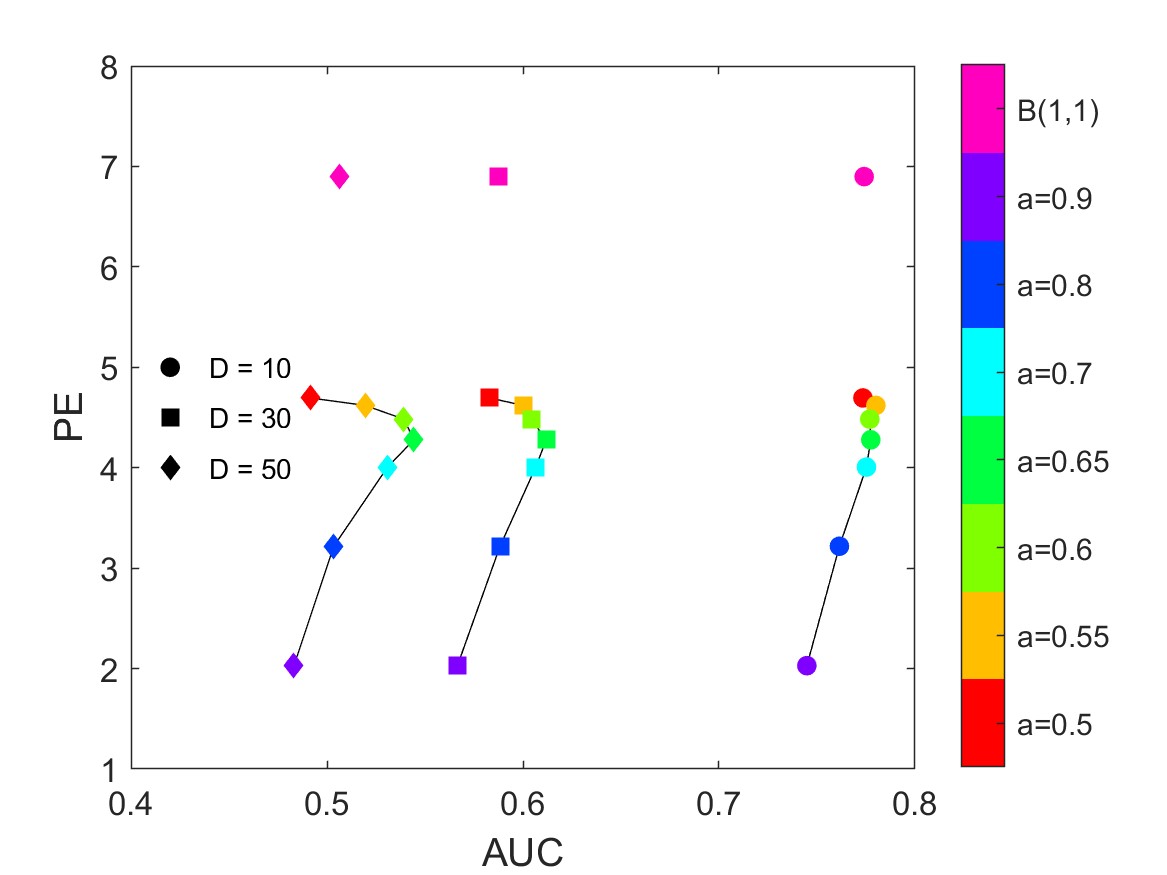}

PE: \hspace{0.3cm} (c) Arcsine \hspace{3cm} (f) Uniform \hspace{1.2cm}
 
\caption{Area under the EAF curve scores (AUC) vs. quantitative measures of chaos for the whole numerical budget $[1,10\:000]$: (a), (d Lyapunov exponent (LE), (b), (e) mutual information (MI) with $\tau=2$, (c), (f) permutation entropy (PE) with $\tau=1$ and $d=5$. Dots, squares and diamonds represent the search space dimension $D=\{10,30,50\}$, while the color code gives the parameter values of the maps. }
\label{fig_measures_all}

\end{figure}

\begin{figure}

\includegraphics[trim = 9mm 3mm 5mm 8mm,clip, width=6.2cm, height=4.5cm]{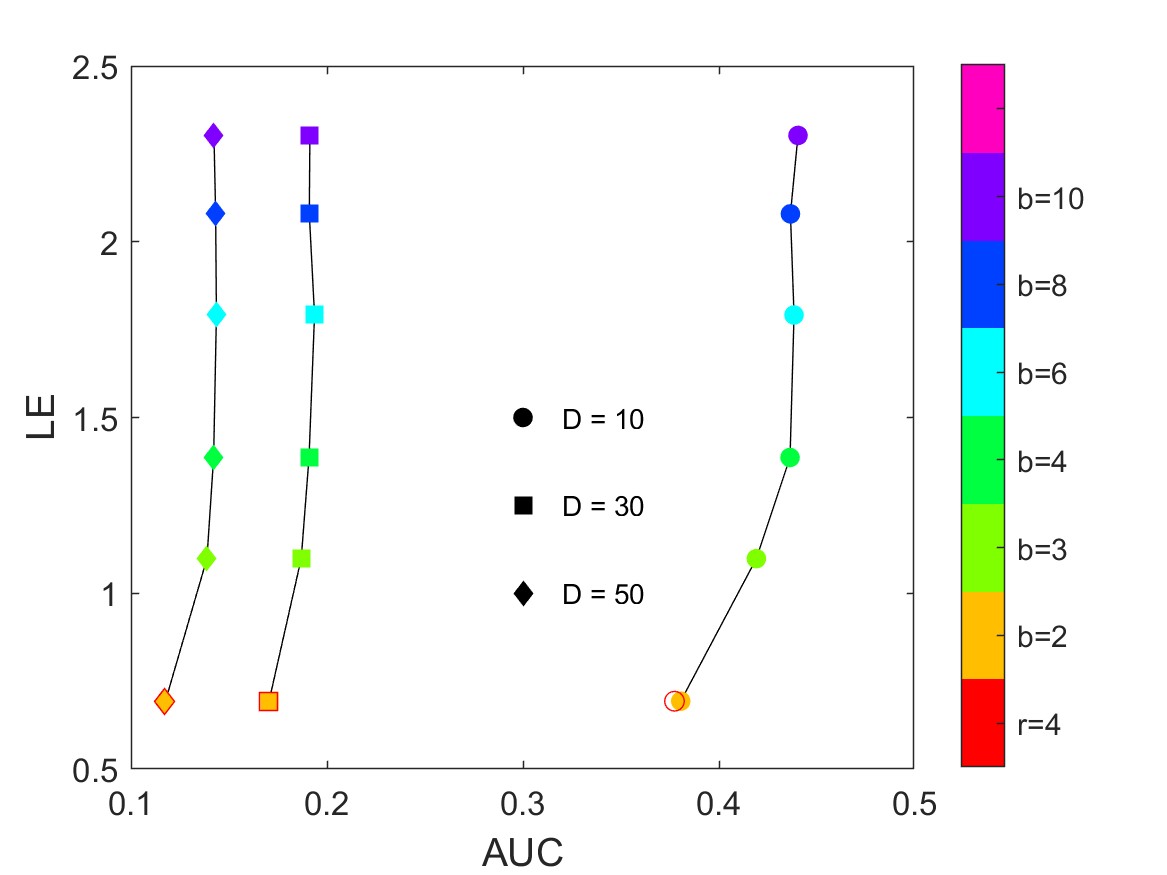}
\includegraphics[trim = 9mm 3mm 5mm 8mm,clip, width=6.2cm, height=4.5cm]{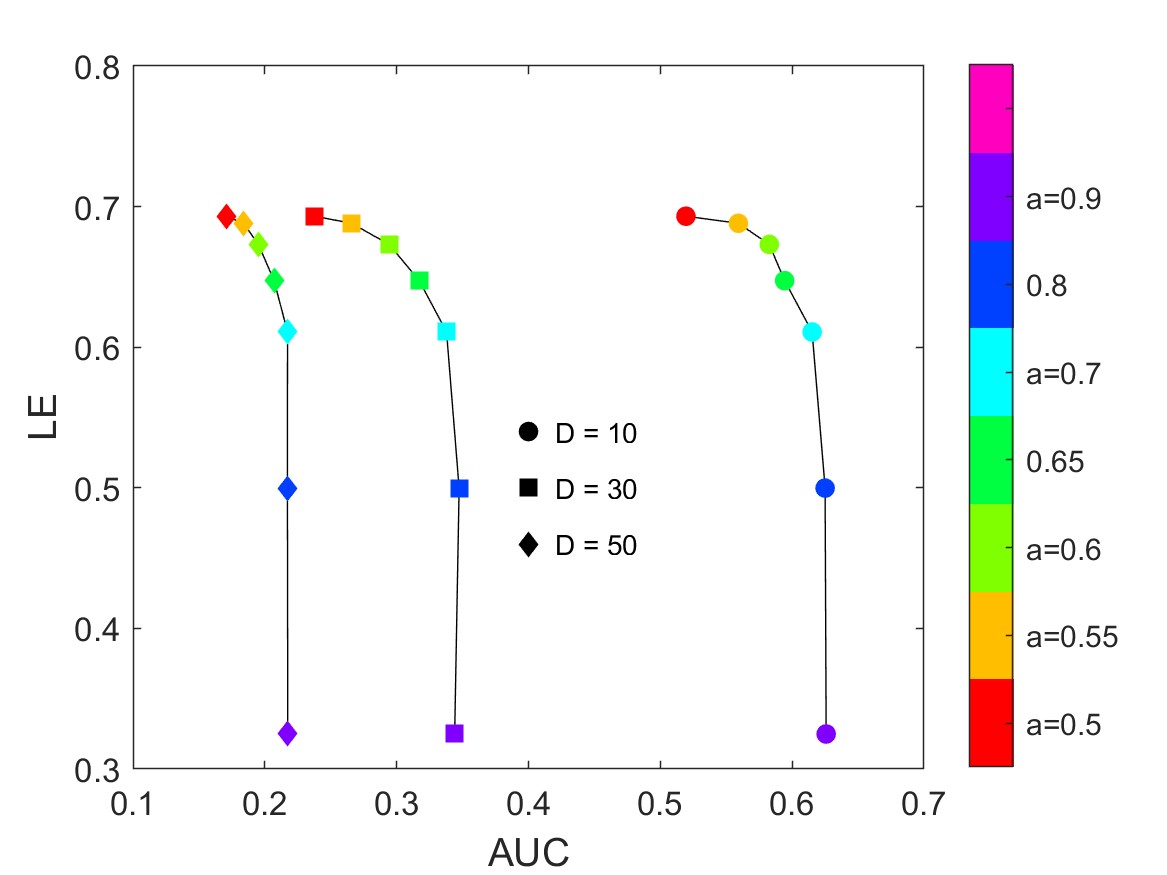}

 LE: \hspace{0.3cm} (a) Arcsine \hspace{3cm} (d) Uniform \hspace{1.2cm}

\includegraphics[trim = 9mm 3mm 5mm 8mm,clip, width=6.2cm, height=4.5cm]{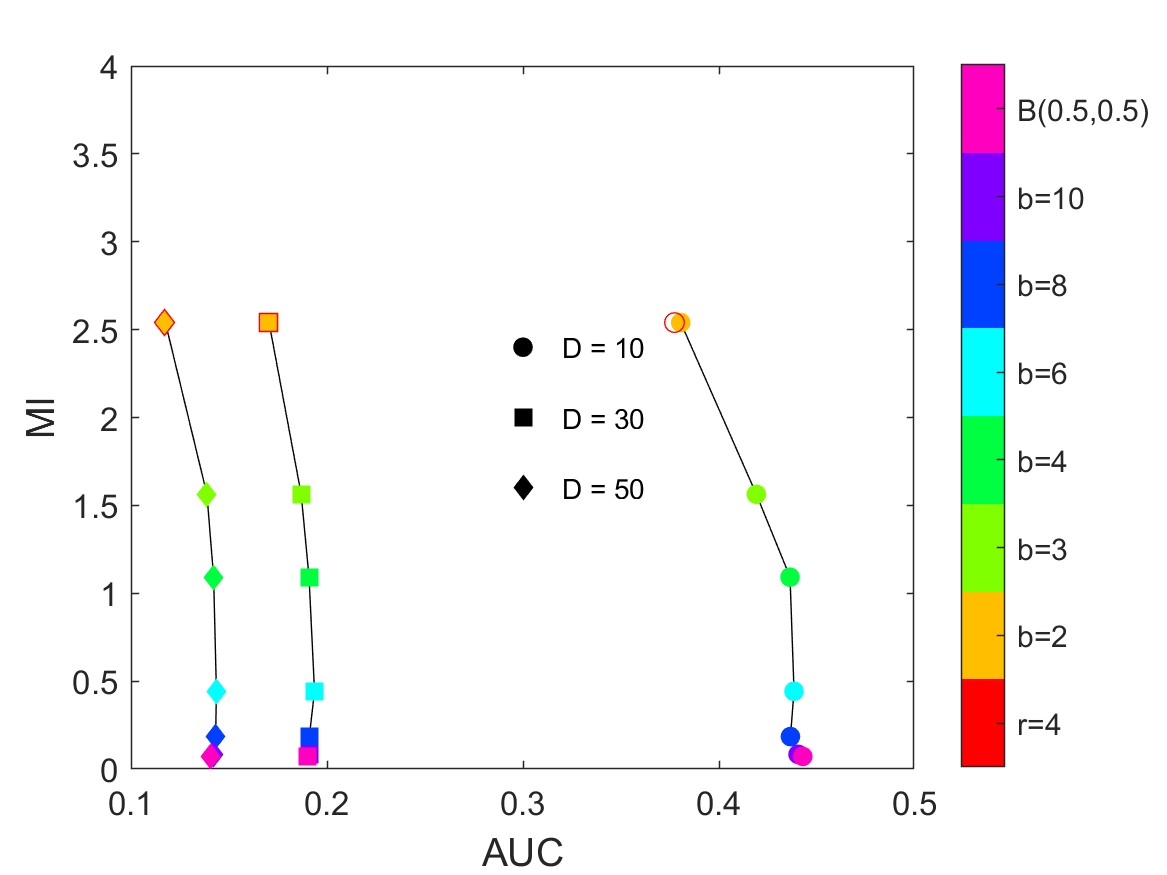}
\includegraphics[trim = 9mm 3mm 5mm 8mm,clip, width=6.2cm, height=4.5cm]{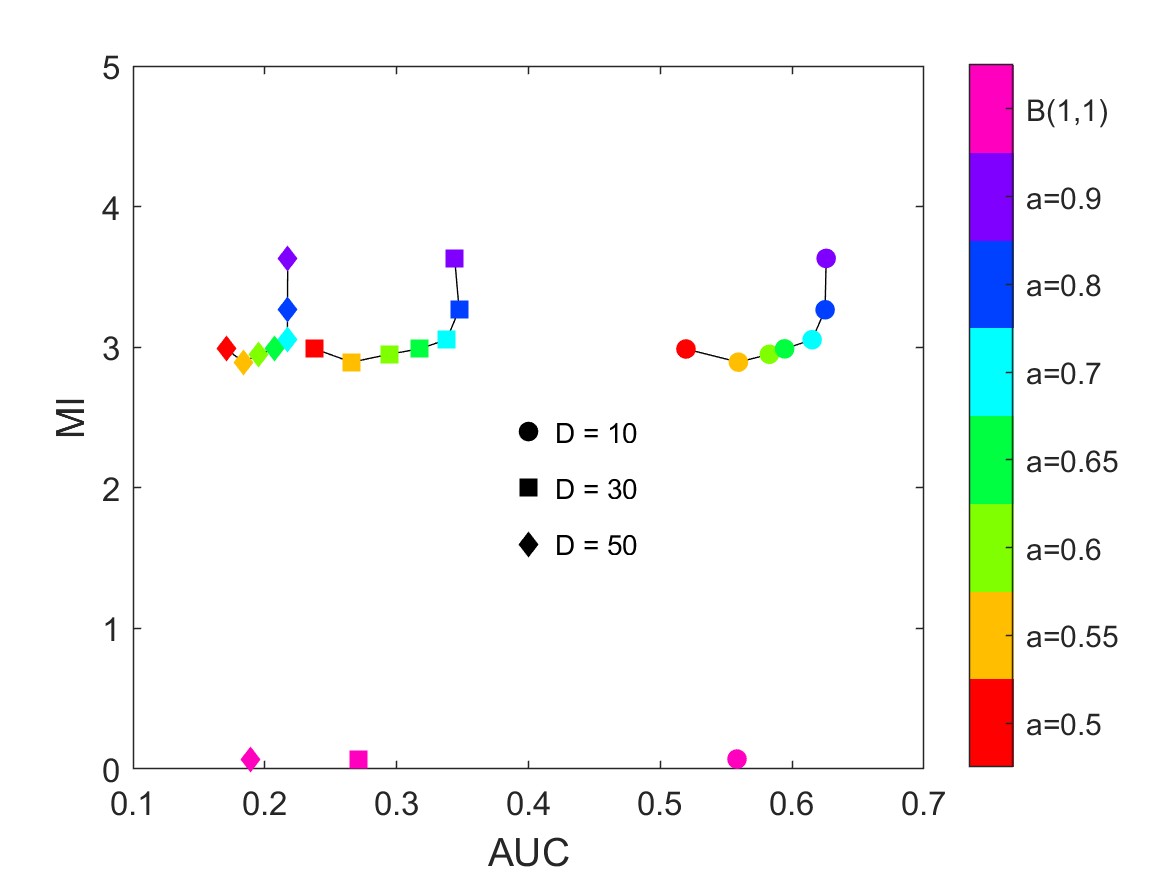}

MI: \hspace{0.3cm} (b) Arcsine \hspace{3cm} (e) Uniform \hspace{1.2cm}

\includegraphics[trim = 9mm 3mm 5mm 8mm,clip, width=6.2cm, height=4.5cm]{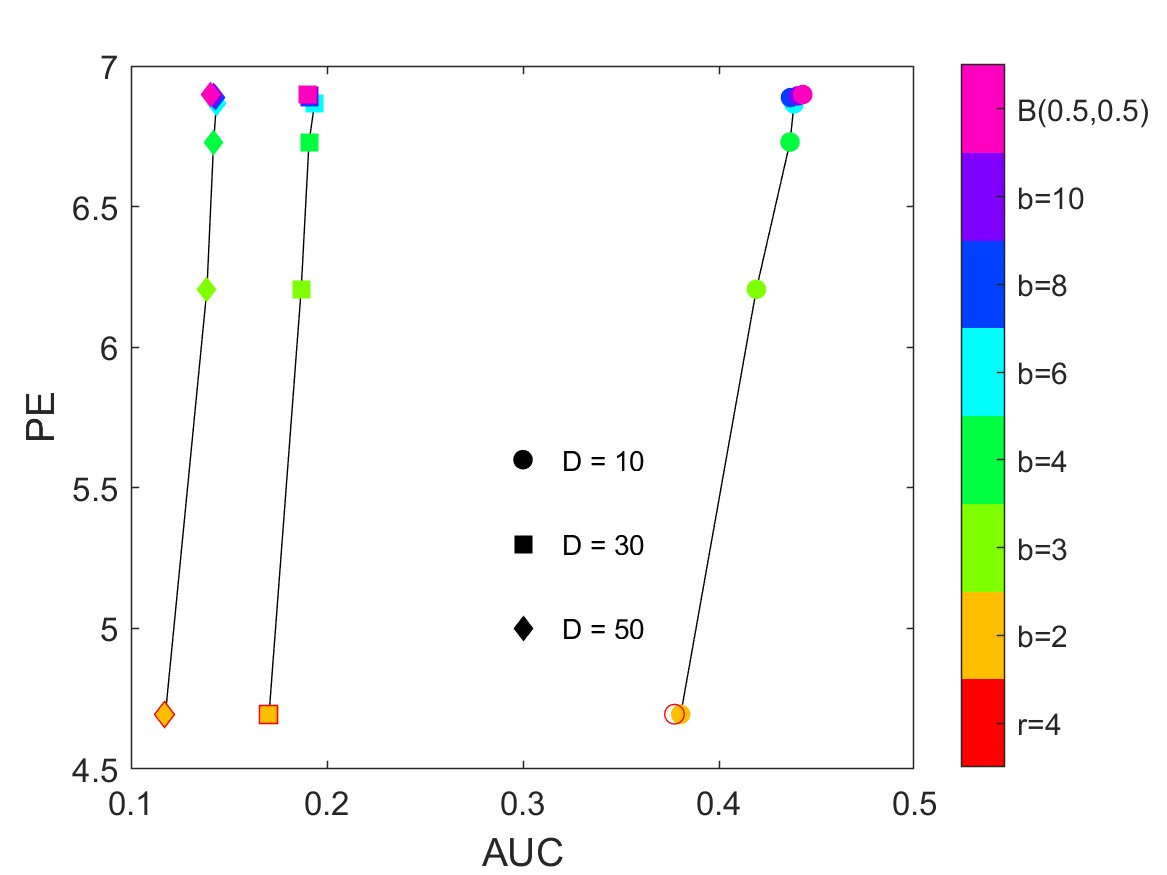}
\includegraphics[trim = 9mm 3mm 5mm 8mm,clip, width=6.2cm, height=4.5cm]{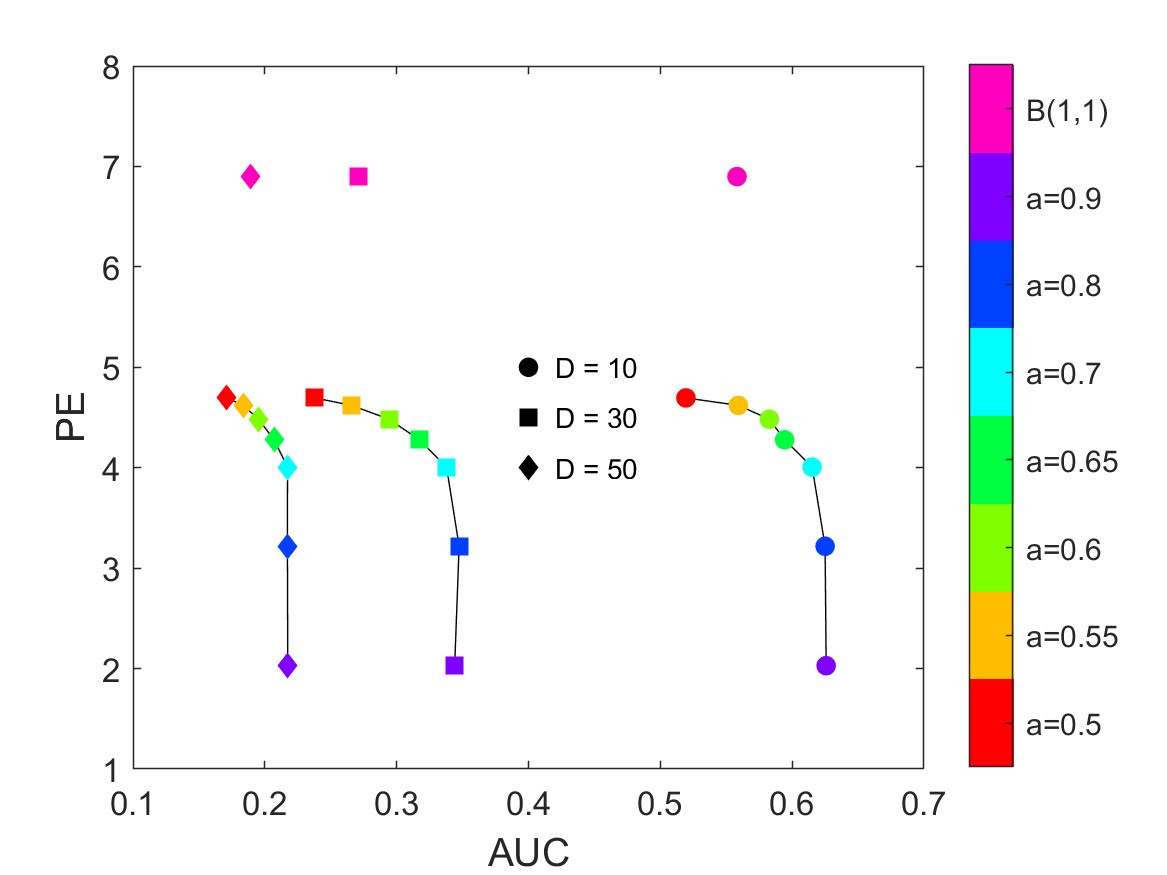}

PE: \hspace{0.3cm} (c) Arcsine \hspace{3cm} (f) Uniform \hspace{1.2cm}
 
\caption{Area under the EAF curve scores (AUC) vs. quantitative measures of chaos for a limited numerical budget range $[20,200]$: (a), (d) Lyapunov exponent (LE), (b), (e) mutual information (MI) with $\tau=2$, (c) (f) permutation entropy (PE) $\tau=1$ and $d=5$. Dots, squares and diamonds represent the search space dimension $D=\{10,30,50\}$, while the color code gives the parameter values of the maps.}
\label{fig_measures}

\end{figure}
\subsection{AUCs and Chaotic Dynamics Measures} \label{sec:chao_auc}

The performance differences for varying the parameters of the chaotic maps observed in $\Delta \: \text{Fraction}$ (Fig.~\ref{fig_anytime_comp}) suggest the question of quantified relationships between chaotic dynamics and PSO performance. Figs. \ref{fig_measures_all} and \ref{fig_measures} show the area under the EAF curve, AUC, which is a measure of the anytime performance over a given budget~\cite{lo24,niki24}, vs. three quantitative measures of chaotic dynamics, the Lyapunov exponent (LE), the mutual information (MI),  and the permutation entropy (PE), see Sec.~\ref{sec:temp_diff}. We discuss two settings. Fig.~\ref{fig_measures_all} shows the AUC over the whole numerical budget of $10\:000$ function evaluations (thus the budget range $[1,10\:000]$), while Fig.~\ref{fig_measures} gives the AUC over the numerical budgets between 20 and 200 function evaluations (a limited range $[20,200]$). It thus accounts for the effect which such a budget restriction would have on anytime performance. Dots, squares and diamonds represent the search space dimension $D=\{10,30,50\}$, while the color code at the right hand side of the figures gives the parameter values $b$ and $a$ of the maps. Note that for AUC vs. LE, Figs. \ref{fig_measures_all}(a) and (d), and \ref{fig_measures}(a) and (d), the value for random sequences is absent, as LE is undefined (infinite) for random, by definition.  

An overall finding is the (expected) inverse relationship between the AUC and the search space dimension $D$: the AUC is largest for $D=10$, smaller for $D=30$, and smallest for $D=50$. Another result is that the anytime performance of uniform distribution is better than for arcsine distribution, which has already been noticed by the EAFs (see Sec.~\ref{sec:res_any}). Tab. \ref{tab:boots} reports the mean AUC together with the lower and upper 95\% bootstrap confidence intervals using 5000 bootstrap samples for the budget range $[1,10\:000]$, see the supplemental material for the range $[20,200]$. For $D=\{30.50\}$ there is a considerable gap between the confidence intervals of the best mean AUC values obtained for arcsine and uniform values. The results also confirm previous findings on random and chaos comparing distributions~\cite{rich24}. For both distributions, random is among the best, but neither for the whole budget $[1,10\:000]$ nor for the limited budget $[20,200]$ the differences are statistically significant according to the experimental setup of 15 independent runs on 3 instances of 24 BBOB benchmark functions. Nonetheless, the results also show that there are  quantifiable differences among different parameter values of the chaotic maps, and also with respect to random. In particular, for the dimensions $D=\{30,50\}$ intermediate values of the parameter $b$ and $a$ seem to be advantageous, and for uniformly distributed values, the margin between the best chaotic values and second-best random is largest.

\begin{table}[htbp]
\centering
\caption{Area under the EAF curves (AUC) quantification for the whole numerical budget $[1,10\:000]$:  means (M),  and lower and upper 95\% bootstrap confidence intervals (CI) using 5000 bootstrap samples over 15 PSO runs with $24 \times 3=72$ BBOB function–instance combinations. Upper half: arcsine distributed values by the Schröder map, Eq.~\eqref{eq:schr} and $\mathcal{B}(z,0.5,0.5)$. Lower half: uniform distributed values by the asymmetric tent map, Eq.~\eqref{eq:atent} and $\mathcal{B}(z,1,1)$. Last lines labeled by R: random values. Bold print: best AUC for each $D$ and distribution.   }
\label{tab:boots}
\scalebox{0.75}{
\begin{tabular}{c| cc | cc |cc}
\toprule
 & \multicolumn{2}{c}{$D=10$} & \multicolumn{2}{c}{$D=30$} & \multicolumn{2}{c}{$D=50$} \\
\cmidrule(lr){2-3} \cmidrule(lr){4-5} \cmidrule(lr){6-7}
r/b & M & CI & M & CI & M & CI\\
 \midrule
4 &   0.6992    &  [0.6390,    0.7511] &    0.2793    &  [0.2232,    0.3412] &    0.1874    &  [0.1388,    0.2460] \\
2 &   0.7006    &  [0.6404,    0.7529] &    0.2792    &  [0.2211,    0.3448] &    0.1865    &  [0.1366,    0.2452] \\
3 &   0.7240    &  [0.6675,    0.7774] &    0.3259    &  [0.2664,    0.3932] &    0.2067    &  [0.1577,    0.2686] \\
4 &   0.7319    &  [0.6745,    0.7819] &    {\bf0.3696}    &  [0.3102,    0.4337] &    {\bf0.2218}    &  [0.1665,    0.2797] \\
6 &  0.7349    &  [0.6786,    0.7878] &    0.3646    &  [0.3059,    0.4290] &    0.2182    &  [0.1676,    0.2796] \\
8 &   0.7347    &  [0.6775,    0.7844] &    0.3660    &  [0.3025,    0.4284] &    0.2199    &  [0.1677,    0.2828] \\
10 &    0.7290    &  [0.6700,    0.7790] &    0.3585    &  [0.2961,    0.4230] &    0.2195    &  [0.1682,    0.2800] \\
R &   {\bf0.7361}    &  [0.6808,    0.7872] &    0.3629    &  [0.3027,    0.4260] &    0.2184    &  [0.1653,    0.2785] \\
\midrule
a & M & CI & M & CI & M & CI\\
 \midrule
0.5 & 0.7734     &  [0.7189,    0.8201] &    0.5829     &  [0.5205,    0.6456] &    0.4466     &  [0.3886,    0.5074] \\
0.55 &    {\bf0.7801}     &  [0.7252,    0.8265] &    0.6000     &  [0.5406,    0.6626] &    0.4728     &  [0.4135,    0.5332] \\
0.6  &  0.7769     &  [0.7234,    0.8254] &    0.6041     &  [0.5410,    0.6654] &    0.4914     &  [0.4326,    0.5523] \\
0.65 &   0.7774     &  [0.7213,    0.8237] &   {\bf0.6119}     &  [0.5490,    0.6732] &    {\bf0.4962}     &  [0.4356,    0.5571] \\
0.7  &  0.7752     &  [0.7218,    0.8246] &    0.6065     &  [0.5409,    0.6679] &    0.4845     &  [0.4207,    0.5457] \\
0.8  &  0.7614     &  [0.7061,    0.8116] &    0.5883     &  [0.5279,    0.6490] &    0.4597     &  [0.3977,    0.5165] \\
0.9  &  0.7448     &  [0.6889,    0.7965] &    0.5665     &  [0.5012,    0.6317] &    0.4410     &  [0.3840,    0.5016] \\
R  &  0.7741     &  [0.7166,    0.8206] &    0.5874     &  [0.5251,    0.6483] &    0.4610     &  [0.4014,    0.5208] \\
\bottomrule
\end{tabular}
}
\end{table}

Regarding the quantitative measures of chaotic dynamics, we find that 
within the parameter range considered,  LE and PE rise monotonically with increasing parameters $b$ of the Schröder map, but fall monotonically with increasing parameter $a$ of the asymmetric tent map. 

Across the whole budget range \([1,10\:000]\), uniformly distributed chaotic values generated using map parameters associated with large LEs consistently exhibit rather low AUC. The AUC increases with falling LE before decreasing slightly (or even largely for uniform distribution and $D=\{30,50\}$)  for LE getting even smaller. With arcsine-distributed values, we observe high AUCs for larger LEs and a  decline in AUC as LE decreases, but the effect is weaker than for uniform. For the  budget range \([20,200]\) arcsine values give similar results as for the whole budget, while for uniform values  the decrease in AUC for lower values of LE does not occur. We have a rather straight inverse relationship between LE and AUC.

The results set into a broader context previous findings which showed that metaheuristic performance can be related to LEs, and larger LEs generally appeared to support performance, but gave no robust and reliable results as  the experimental setup did not disentangle distributional and dynamical influences and did also not apply competitive benchmarks~\cite{yang14,yang23}. Given the setup of distributional equivalence, the findings reported here give a more complex picture. At least for PSO, large LEs are not unanimously connected to better performance, but it depends on the distribution, the numerical budget (and possibly also on the selection of benchmarks). Most clearly, this can be seen by the overall result that (for the chaotic maps considered) arcsine values have a much larger LE than uniform values, but uniform values perform substantially better. 

For the whole numerical budget $[1,10\:000]$ and arcsine distribution, larger LEs indeed give (slightly) better performance, but for uniform distribution, an intermediate LE is connected to best results. For the limited budget range $[20,200]$, we get similar results for arcsine distribution, but for uniform distribution higher values of LE perform less good than intermediate ones.  Smaller LEs mean slower divergence of sequences with nearby seeds and slower decorrelation of values, while larger LEs have the opposite effect~\cite{ott93,wern17}. This motivates the hypothesis that large LEs facilitate exploration, leading to better performance~\cite{yang14,yang23}. 

In view of anytime performance over a numerical budget, this hypothesis must be reconsidered.  The results presented in Figs. \ref{fig_measures_all} and \ref{fig_measures} can be interpreted as indicating a diminishing influence of exploration on performance (particularly for larger budgets). Thus, large LEs  are not unanimously related to better performance at all times. PE calculated for $\tau=1$ and $d=5$ largely mirrors these relationships between LE and AUC (the Pearson correlation between LE and PE is 0.8852 for arcsine and 0.9992 for uniform values, see also the supplemental material). Also, for MI, the results support the interpretation given for the relationships between LE and AUC (the Pearson correlation between LE and MI  for $\tau=2$ is -0.9831 for arcsine and -0.9909 for uniform values).  For arcsine distribution smaller values of MI indicate slightly larger values of AUC for both budget ranges and for uniform and the whole budget $[1,10\:000]$. We see the largest AUC for intermediate values of MI, while for the limited budget $[20,200]$ we do not find the rebound for larger MIs.  

Some final remarks concern the relationships between MI, PE, and AUC for random values. As noted earlier, unlike LEs, these quantities can also be computed for random values. We observe that MI gives much lower values for both arcsine and uniform distributions ($\mathcal{B}(z,0.5,0.5)$ and $\mathcal{B}(z,1,1)$), and PE gives much higher values for uniform. For arcsine distributions, random values give PE similar to the Schröder map with $b=\{6,8,10\}$. This means that for the considered chaotic maps, random and chaos can be clearly separated by MI values. However, the results also show that for random there is no clear relationship between MI, PE and AUC, which is mainly due to the observation reported above that distributionally equivalent random and chaotic values show no systematic performance differences.  Even if MI and PE have different values for random and chaos, this difference does not imply performance differences. Further discussions about MI and PE over varying time-lag $\tau$ are given in the supplemental material.

\section{Conclusions}

\subsection{Main Results}
We have considered PSO performance depending on the usage of either random or chaotic values for driving the evolutionary search.  We use an experimental setup with parametrizable chaotic maps, the Schröder map and the asymmetric tent map, which  consistently have values with the same distribution (arcsine and uniform), but varying chaotic dynamics properties. This setup ensures a fair comparison between random and chaotic values with isolating distributional effects (i.e., the likelihood of particular values occurring) from temporal-dependence effects (i.e., the correlations between successive values). Using an anytime performance assessment, we have studied  
empirical attainment functions (EAFs), which quantify the fraction of runs reaching a given target value within a specified budget, measured in runtime by function evaluations.

The results obtained challenge any assumption of  general performance advantages of chaos over randomness in PSO. Thus, previous results that detected a general ''chaotic exploitation instinct``~\cite{moza19} without controlling distributional effects should be treated with some skepticism.  Instead, the results presented here reveal (more or less subtle) performance differences that depend on the underlying distribution of values, benchmark selection, and numerical budget. 

The findings further support previous observations~\cite{noer23,rich24} that the distribution of values is the most influential factor in PSO performance.  Once distributional effects are controlled, an interesting result of the anytime analysis indicates  performance advantages of chaotic over random values at small budgets, which frequently deteriorate for numerical budget becoming larger. A possible explanation might be that early exploration is most effected by the search directions dictated by the nonlinear maps, see also the discussion in Sec.~\ref{sec:temp_diff}, and this effect wears off for numerical budgets increasing.  Thus, the results of anytime performance may also (at least partly) explain the ambiguous results of random vs. chaos reported in the literature~\cite{capo03,gag21,zel22,liman25} as advantages of one over the other may change not only with problems and implementations, but also with budgets and targets.  
 
Another group of results deals with the relationships between the area under the EAF curve (AUC) as a quantification of the PSO anytime performance and chaotic dynamics measures. As measures we considered Lyapunov exponents (LEs), mutual information (MI), and permutation entropy (PE), with the latter two being calculable for both random and chaotic values. With distributional equivalence and over numerical budgets, no simple relationships can be established between chaotic dynamics measures and PSO performance.
While for arcsine distributions large LEs are related to best performance, for uniform distributions rather intermediate LEs are good. Large LEs mean faster divergence of sequences with seeds close to each other and faster decorrelation~\cite{ott93,wern17}. Thus, it has been assumed that these properties of chaos promote exploration~\cite{yang14,yang23}. Although possibly justified for a fixed budget, the results presented in this paper suggest that this view cannot be maintained across varying budgets or over different value distributions. Given the dominant influence of the value distribution on PSO performance, measures of chaotic dynamics can only be meaningfully related to performance within a given distribution, but not across different distributions.

\subsection{Limitations and Outlook}

The main results suggest that chaos cannot be regarded as a kind of general quality indicator for PSO performance. However, the results are subject to some constraints, as they are based on a finite number of numerical experiments. Thus, the question of whether alternative setups with other benchmarks and/or more repetitions might yield different results remains relevant. In addition, while the results demonstrate good scalability across the dimensions considered, it would be interesting to examine whether this behavior remains stable for substantially higher dimensions. Another limitation of the study is the selected PSO configuration. We used a standard inertia-weight form with common parameter values and a given swarm size, see Sec.~\ref{sec:pso}. It might be possible that the overall findings may vary over the swarm size, and do not transfer unchanged to alternative PSO variants, for instance with other inertia-weight settings and velocity clamping, or more complex PSO implementations (see \cite{hay26,pio23} for an overview). Also this is a topic of future work.

Although such additional (and potentially complementary) results may provide further insights, the main conclusions of this paper should remain unaffected. Our aim is not to establish evidence that any particular setup of random or chaotic values outperforms others for a given group of optimization problems. Rather, we aimed at demonstrating that there are no intrinsic performance differences between random and chaotic sequences if distributional equivalence is observed. If such general differences existed, they should manifest themselves consistently, regardless of the particular collection of benchmark functions considered and the implementation details selected.

As the search dynamics of PSO depends directly on the distributional and temporal properties of the used values,  the effects of taking random or chaotic quantities can be studied in a particularly straightforward and unambiguous way. For other metaheuristics, the random-like influences are arguably more subtle and less discernible, as for instance in mutation or selection. However, it remains to be assumed that these effects still occur (at least in principle) in a similar way, which is a topic of future work.   Also, the results about relationships between chaotic dynamics measures and performance might be a good starting point for work that
enables design and selection of chaotic maps particularly suitable for driving metaheuristics.

\section*{Data and code availability}
 The program code and data to replicate the results 
 can be found at the data repository \url{https://github.com/HendrikRichterLeipzig/PSO_anytime}.

\section*{Statement of Contributions}
In this work, generative artificial intelligence (ChatGPT, OpenAI, GPT-6, and Perplexity.ai) was used for literature search,  grammar and language editing to improve the readability and flow of the text, and to review a first draft. All outputs and sources were checked for accuracy and meaning. The authors maintain full responsibility for the originality, accuracy, and integrity of the final work.

\bibliographystyle{plain}

\bibliography{example}

\section*{Supplemental Material}
\subsection*{S1: Equivalence of Chaotic Maps}

For $b=2$ and $r=4$, and the interval $ x \in [0,1]$, the Schröder map 
\begin{equation*}z(k+1)=\sin^2\left(b \cdot \arcsin{\sqrt{x(k)}} \right) \end{equation*}
and the Logistic map \begin{equation*}z(k+1)=rz(k)(1-z(k))  \end{equation*} are identical:

\noindent Let
\[
\theta = \arcsin(\sqrt{x}) \quad \Rightarrow \quad \sin\theta = \sqrt{x}, \ \theta \in \left[0,\frac{\pi}{2}\right]
\]
and obtain
\[
\cos\theta = \sqrt{1 - \sin^2\theta} = \sqrt{1 - x}.
\]
Now find
\[
\sin(2\theta) = 2\sin\theta\cos\theta = 2\sqrt{x}\sqrt{1-x}
\]
and square
\[
\sin^2(2\theta) = \left(2\sqrt{x}\sqrt{1-x}\right)^2 = 4x(1-x).
\]
As $\theta = \arcsin(\sqrt{x})$, 
it follows
\[
4x(1-x) = \sin^2\!\bigl(2\arcsin(\sqrt{x})\bigr) \quad \text{for } x \in [0,1].
\]

\subsection*{S2: MI and PE Dependency from $\tau$}

\begin{figure}[htb]

\includegraphics[trim = 5mm 3mm 10mm 8mm,clip, width=6.2cm, height=4.5cm]{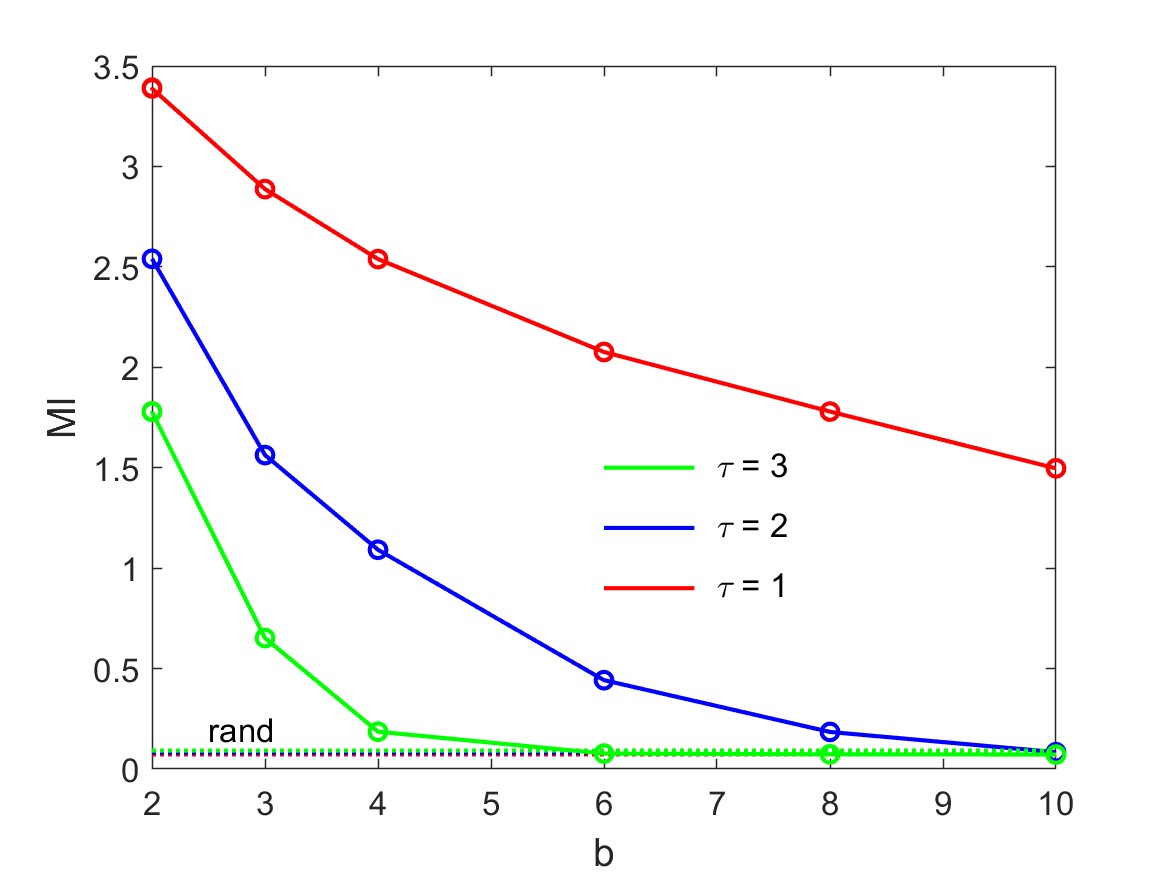}
\includegraphics[trim = 5mm 3mm 10mm 8mm,clip, width=6.2cm, height=4.5cm]{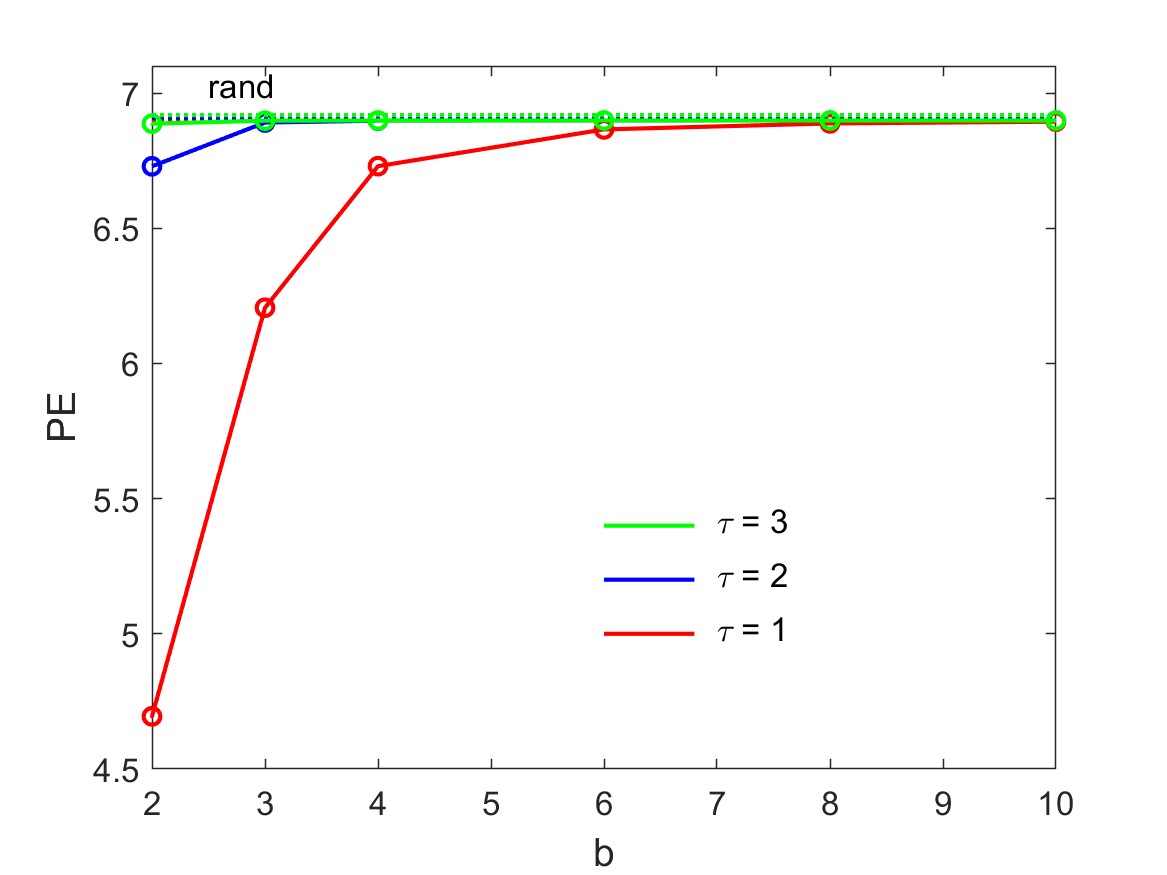}

Arcsine: \hspace{0.3cm} (a) MI  \hspace{2.2cm} (d) PE \hspace{1.3cm} 

\includegraphics[trim = 5mm 3mm 10mm 8mm,clip, width=6.2cm, height=4.5cm]{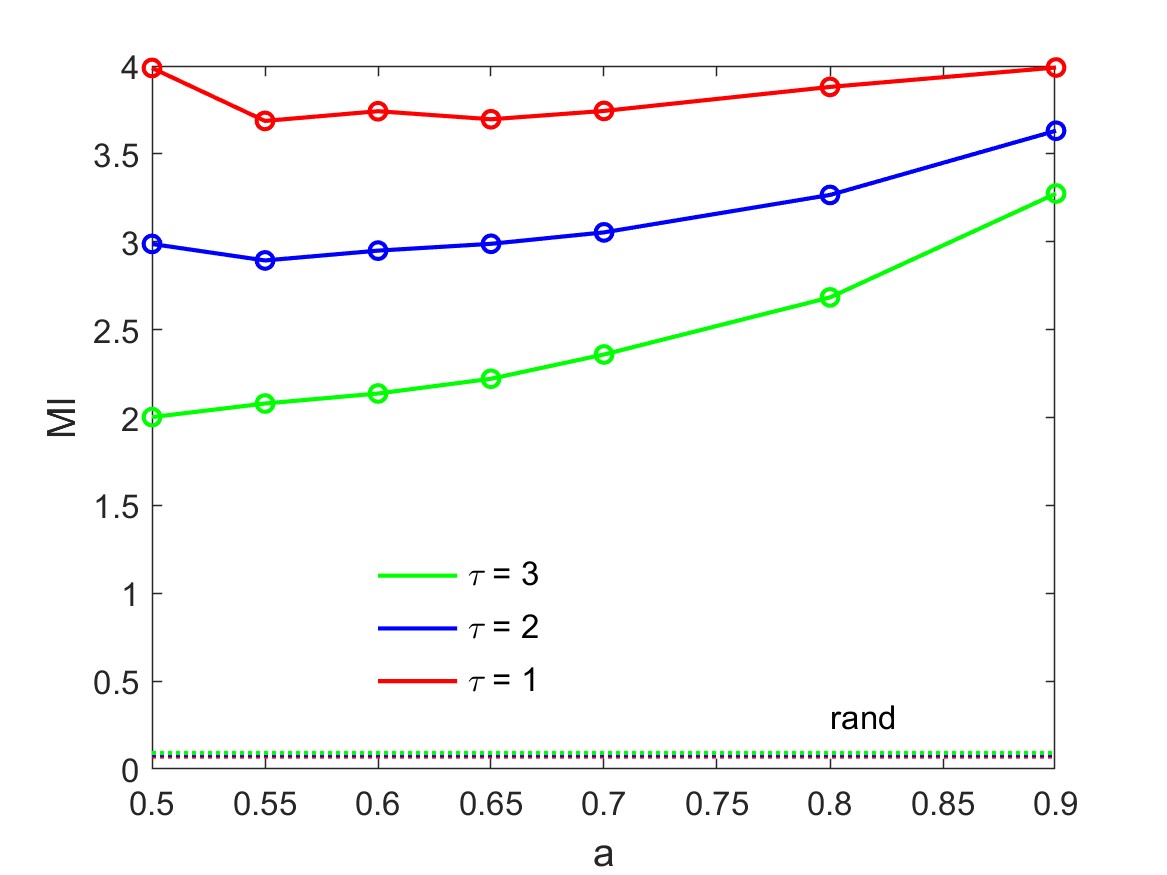}
\includegraphics[trim = 5mm 3mm 10mm 8mm,clip, width=6.2cm, height=4.5cm]{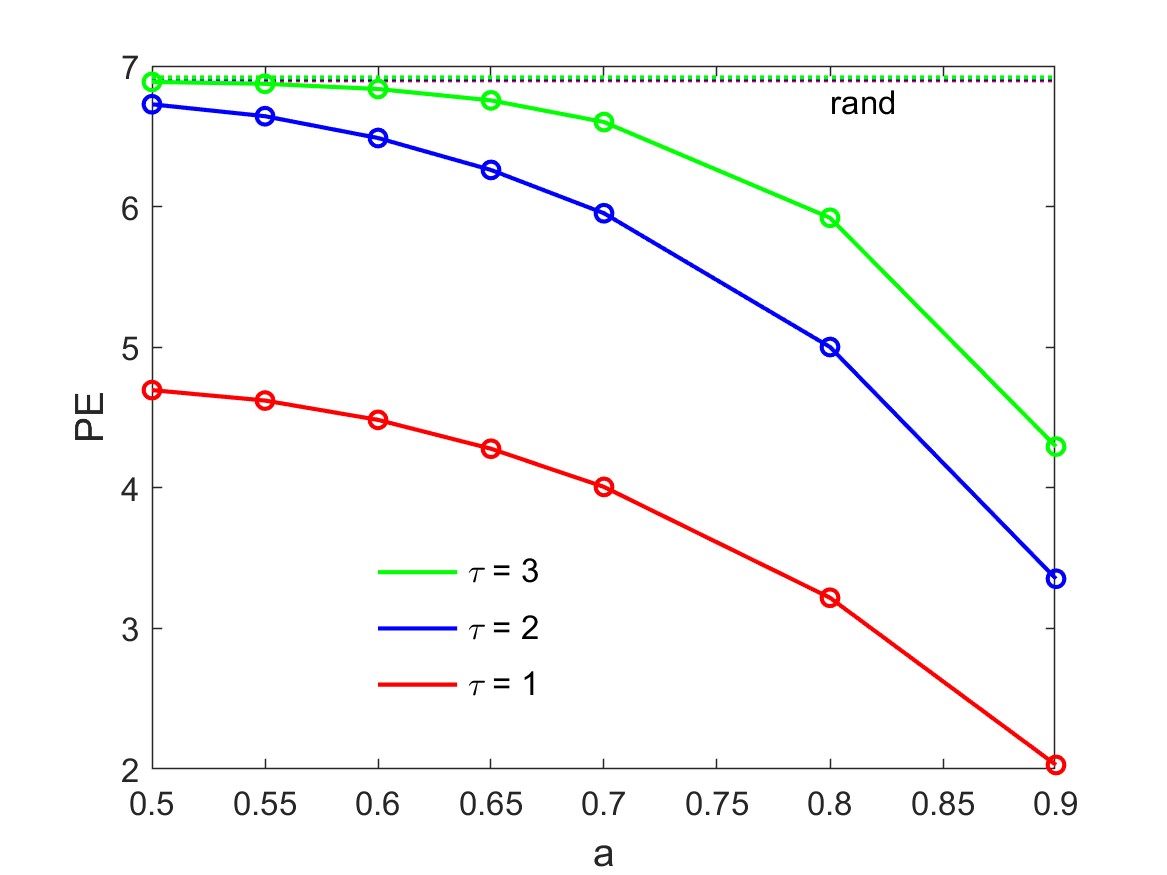}

Uniform: \hspace{0.3cm} (b) MI  \hspace{2.2cm} (e) PE \hspace{1.3cm}

\caption{ MI and PE depending on $\tau$}
\label{fig_mi_pe}

\end{figure}

Fig.~\ref{fig_mi_pe} shows two chaotic dynamics measures, the  mutual information (MI) and  the permutation entropy (PE), over the parameter $b$ of the Schröder map, Eq. (1), and the parameter $a$ of the asymmetric tent map, Eq. (2), for three different time lags $\tau$. Furthermore, the value for random values with the same distribution (arcsine with $\mathcal{B}(z,0.5,0.5)$ and uniform with $\mathcal{B}(z,1,1)$) are shown as dotted line. We observe that except for PE, the Schröder map with $b>2$ and $\tau=\{2,3\}$, MI and PE for chaos are clearly separated for varying $\tau$, while for random MI and PE are almost the same for all $\tau$. In other words, for chaos, MI and PE are sensitive to the time lag $\tau$. We also see that for the Schröder map  (arcsine distribution), MI falls with increasing $b$, while PE rises with $b$. For the asymmetric tent map (uniform distribution), we find  opposite 
relations, see also the correlations between the chaotic dynamics measures in Table 1. We can further see that with increasing $\tau$ MI as well as PE become closer to the values for random. This means while for $\tau=1$ it is almost always possible to distinguish chaos from random, for $\tau=3$, this is frequently not possible.

\begin{table}[htbp]
\centering
\caption{Correlations between LE, MI and PE depending on $\tau$ by Pearson coefficient $r$.}
\label{tab:results}
\begin{tabular}{c| cc | cc |cc}
\toprule
 & \multicolumn{2}{c}{$r(\text{LE,MI})$} & \multicolumn{2}{c}{$r(\text{LE,PE})$} & \multicolumn{2}{c}{$r(\text{MI,PE})$} \\
\cmidrule(lr){2-3} \cmidrule(lr){4-5} \cmidrule(lr){6-7}
$\tau$ & arcsine & uniform & arcsine & uniform & arcsine & uniform \\
\midrule
1 & -0.9997 & -0.5720 & 0.8852 & 0.9992 & -0.8948 & -0.5486 \\
2 & -0.9831 & -0.9909  & 0.7988 & 0.9997 & -0.8827 & -0.9897 \\
3 & -0.8887 &  -0.9982 & 0.8030 & 0.9864 & -0.9766 & -0.9815 \\
\bottomrule
\end{tabular}
\end{table}

Table 2 gives the Pearson correlations between LE, MI and PE for $\tau=\{1,2,3\}$. Note that only MI and PE depend on $\tau$. In other words, we correlate a constant LE to varying MI and PE. We see that generally, there is a rather high correlation between all values, which can be interpreted as the chaotic dynamics quantities scale well with each other for varying parameter values of the chaotic maps. Also note that this implies that the three measures account for the same underlying mechanism of chaos producing random-like values. They are therefore not independent in terms of the properties they quantify.

\subsection*{S3: $\Delta \: \text{Fraction}$ for $D=10$ and $D=50$}

\begin{figure}[tbh]

\includegraphics[trim = 5mm 3mm 10mm 8mm,clip, width=6.2cm, height=4.5cm]{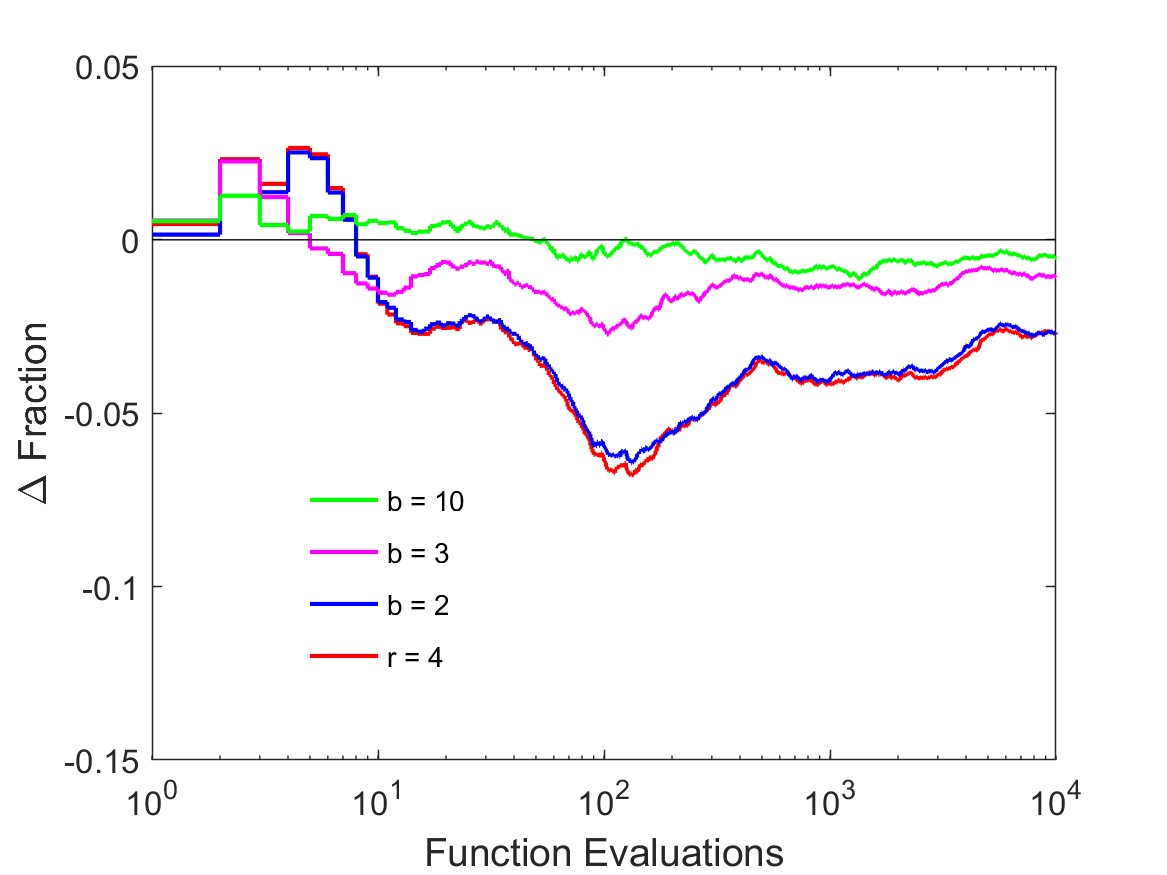}
\includegraphics[trim = 5mm 3mm 10mm 8mm,clip, width=6.2cm, height=4.5cm]{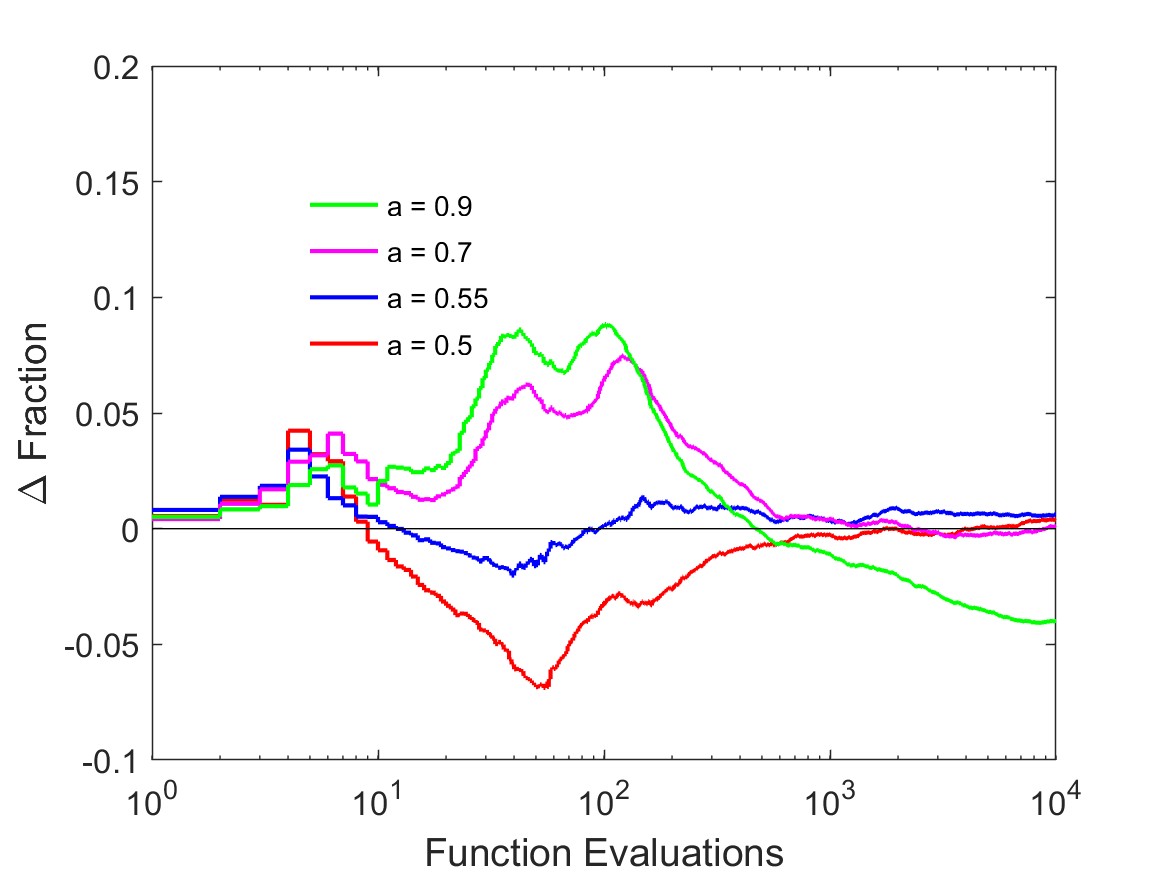}

BBOB 1-24: \hspace{0.3cm} (a) Arcsine  \hspace{2.2cm} (d) Uniform \hspace{1.3cm} 

\includegraphics[trim = 5mm 3mm 10mm 8mm,clip, width=6.2cm, height=4.5cm]{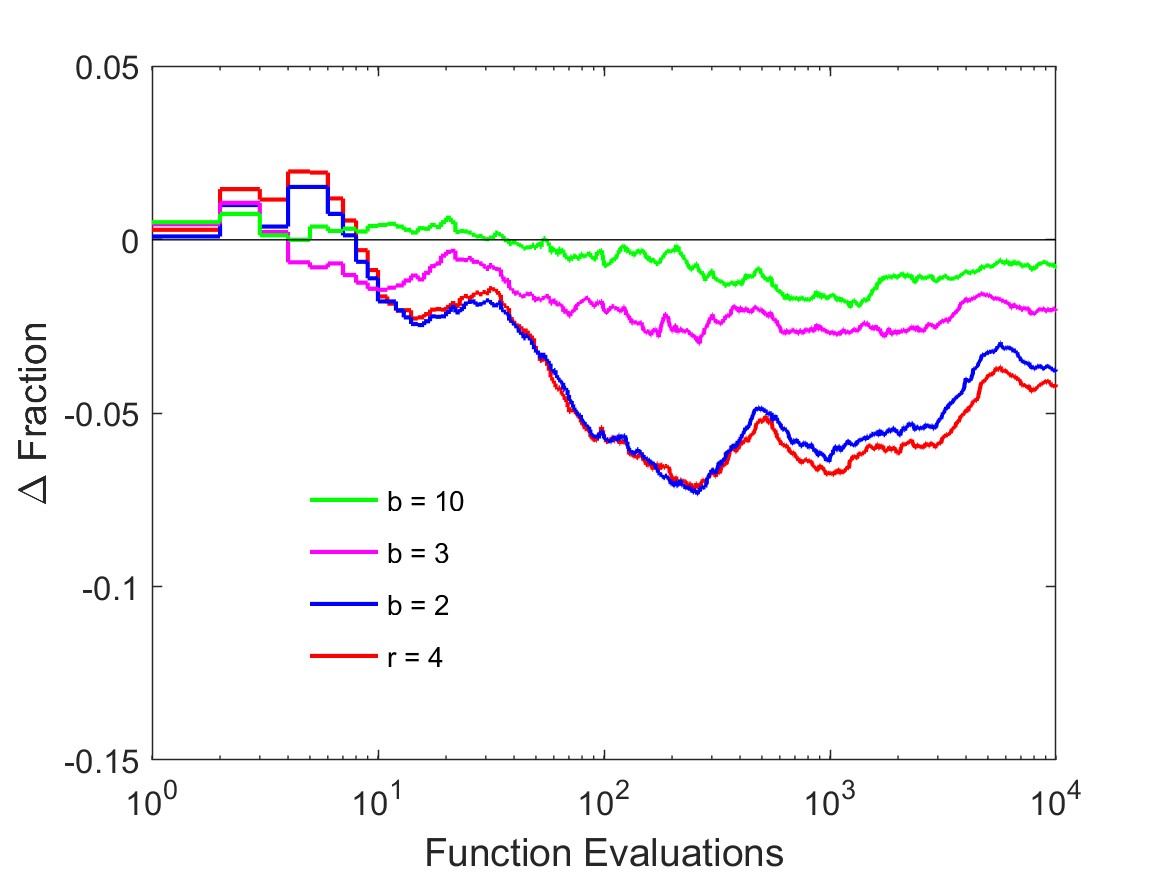}
\includegraphics[trim = 5mm 3mm 10mm 8mm,clip, width=6.2cm, height=4.5cm]{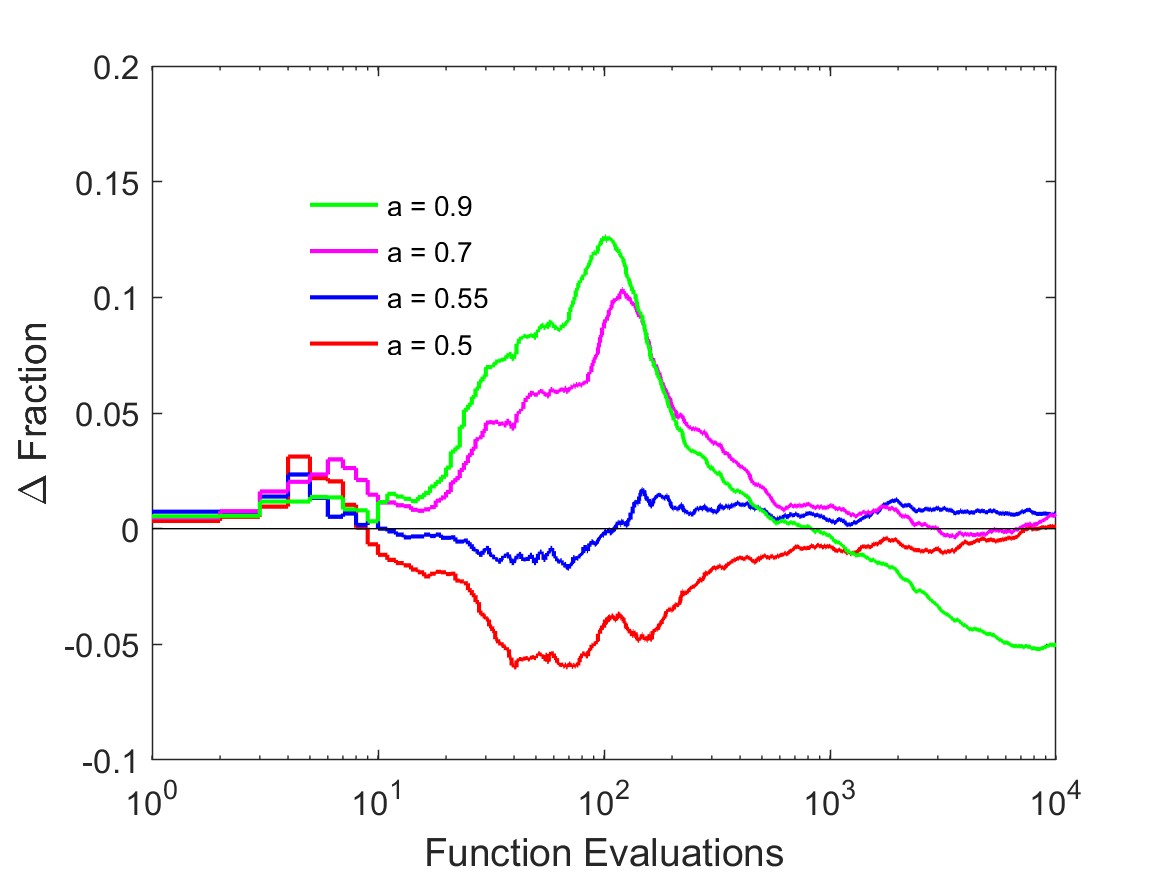}

 BBOB 1-14: \hspace{0.3cm} (b) Arcsine  \hspace{2.2cm} (e) Uniform \hspace{1.3cm}

\includegraphics[trim =7mm 3mm 10mm 8mm,clip, width=6.2cm, height=4.5cm]{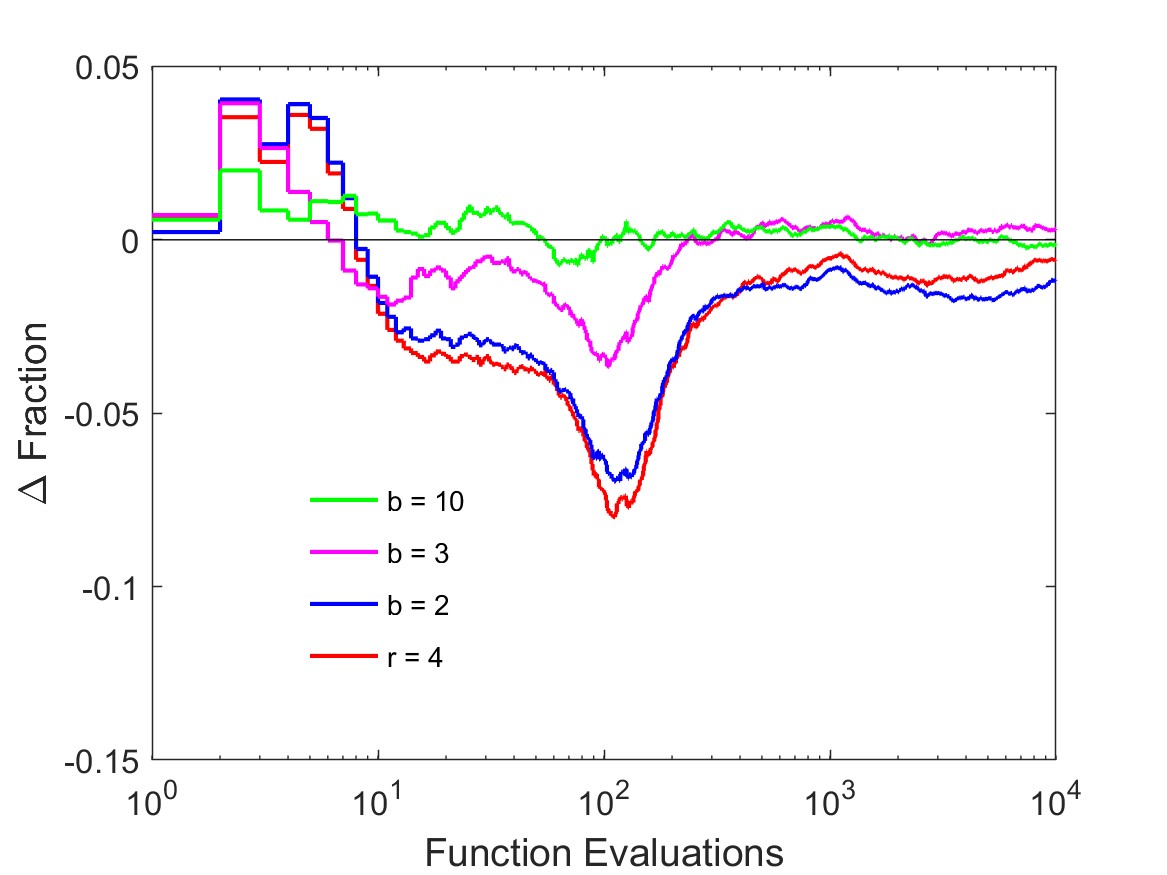}
\includegraphics[trim = 5mm 3mm 10mm 8mm,clip, width=6.2cm, height=4.5cm]{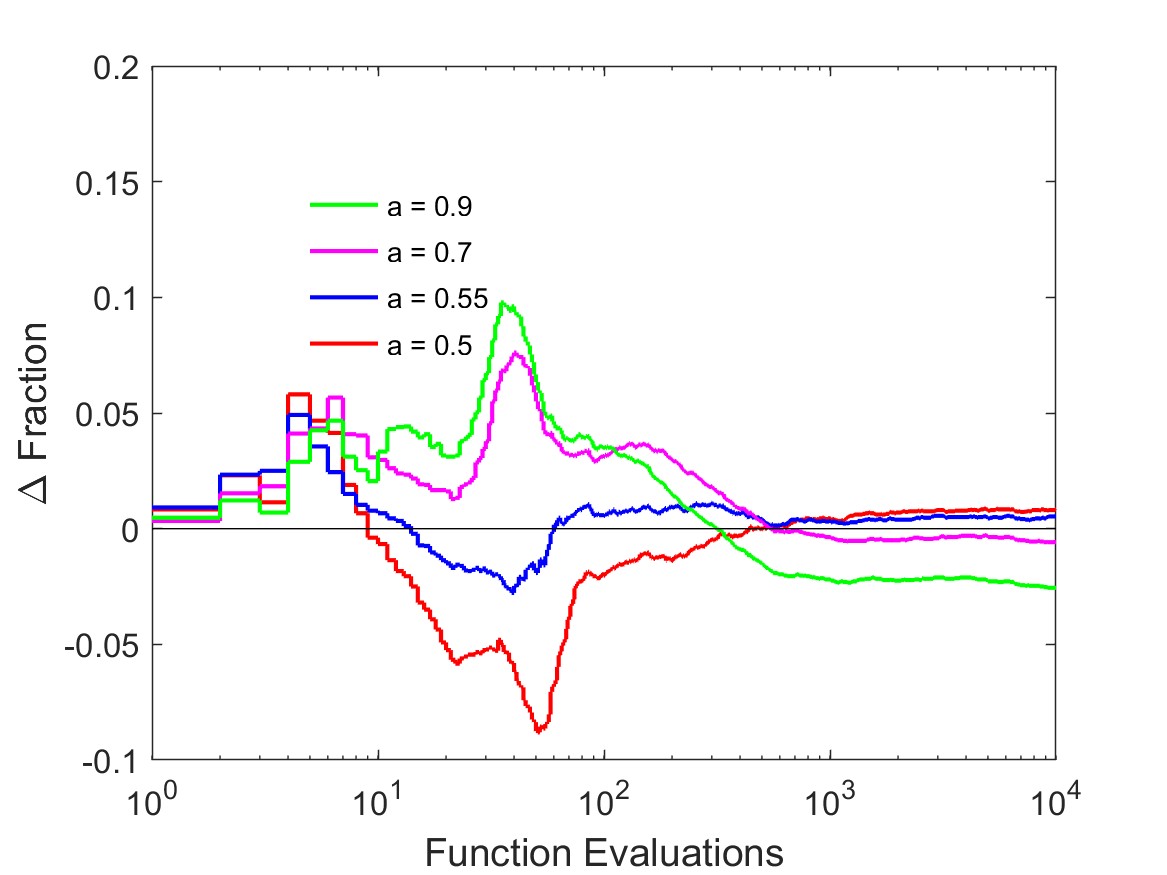}

BBOB 15-24: \hspace{0.2cm} (c) Arcsine  \hspace{2.2cm} (f) Uniform \hspace{1.3cm}
 
\caption{ $\Delta \: \text{Fraction} = \text{EAF}_{chao}-\text{EAF}_{rand}$. Search space dimension $D=10$.}
\label{fig_anytime_comp_10}

\end{figure}

\begin{figure}[tbh]

\includegraphics[trim = 5mm 3mm 10mm 8mm,clip, width=6.2cm, height=4.5cm]{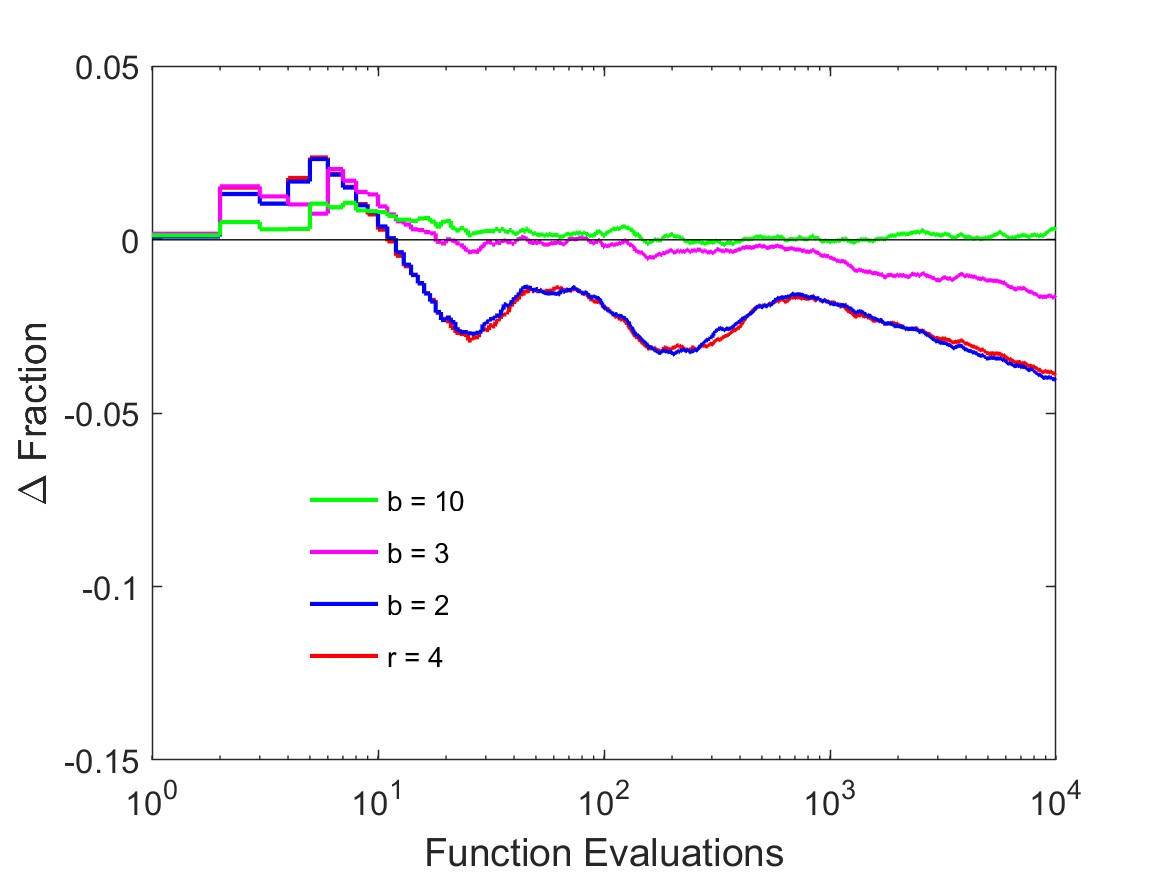}
\includegraphics[trim = 5mm 3mm 10mm 8mm,clip, width=6.2cm, height=4.5cm]{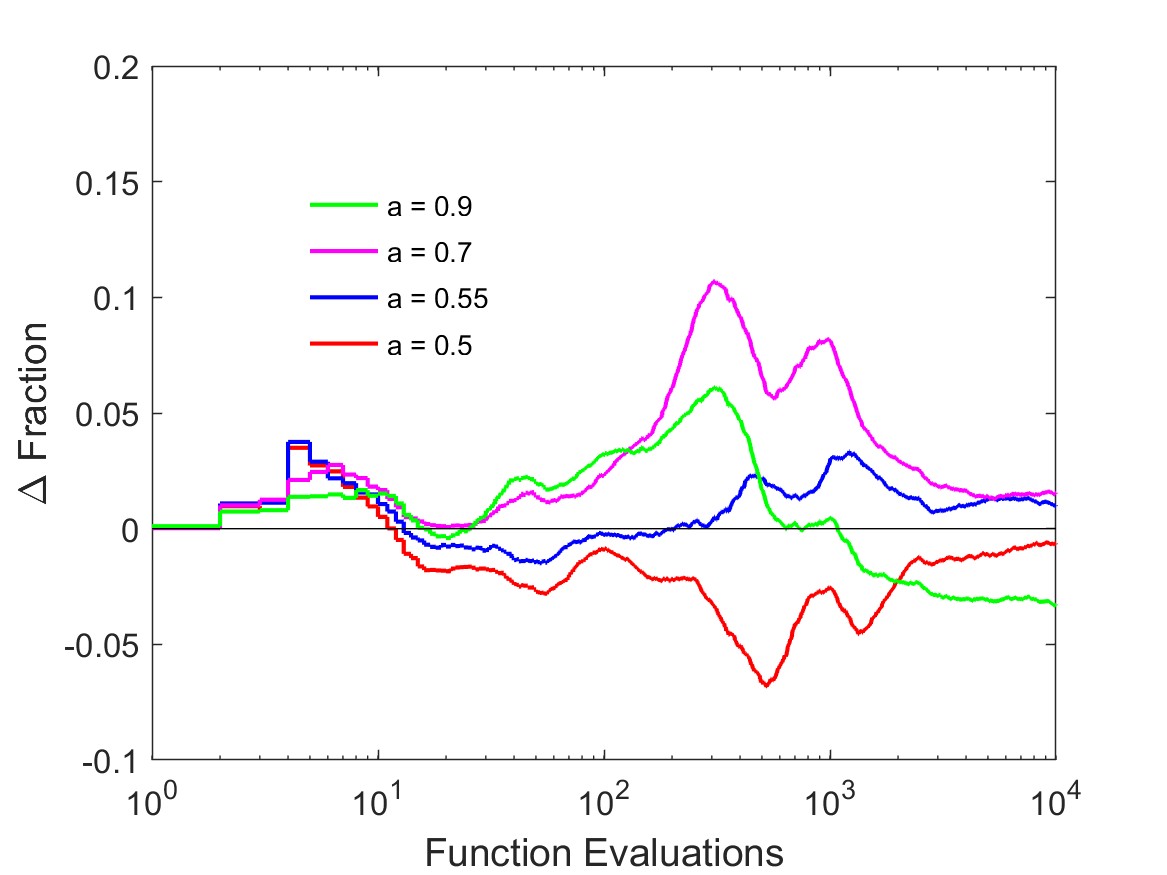}

BBOB 1-24: \hspace{0.3cm} (a) Arcsine  \hspace{2.2cm} (d) Uniform \hspace{1.3cm} 

\includegraphics[trim = 5mm 3mm 10mm 8mm,clip, width=6.2cm, height=4.5cm]{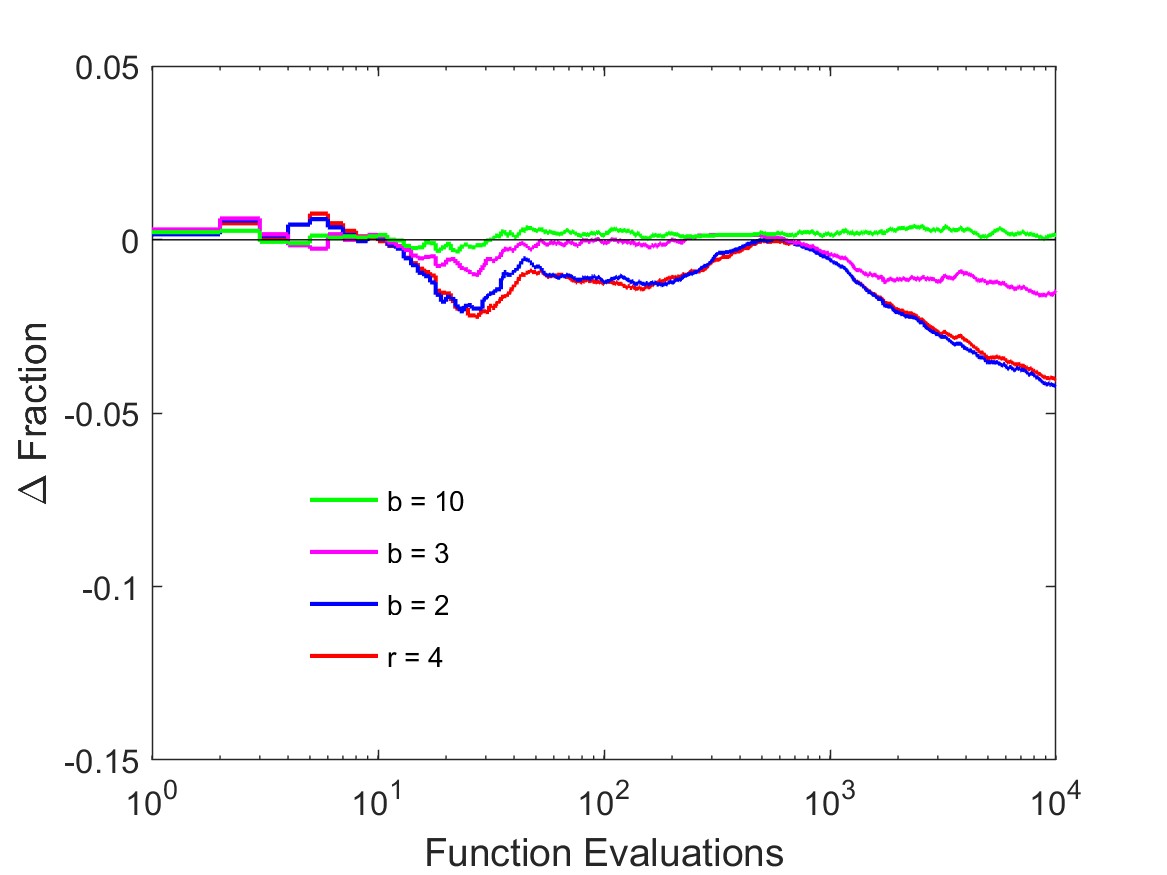}
\includegraphics[trim = 5mm 3mm 10mm 8mm,clip, width=6.2cm, height=4.5cm]{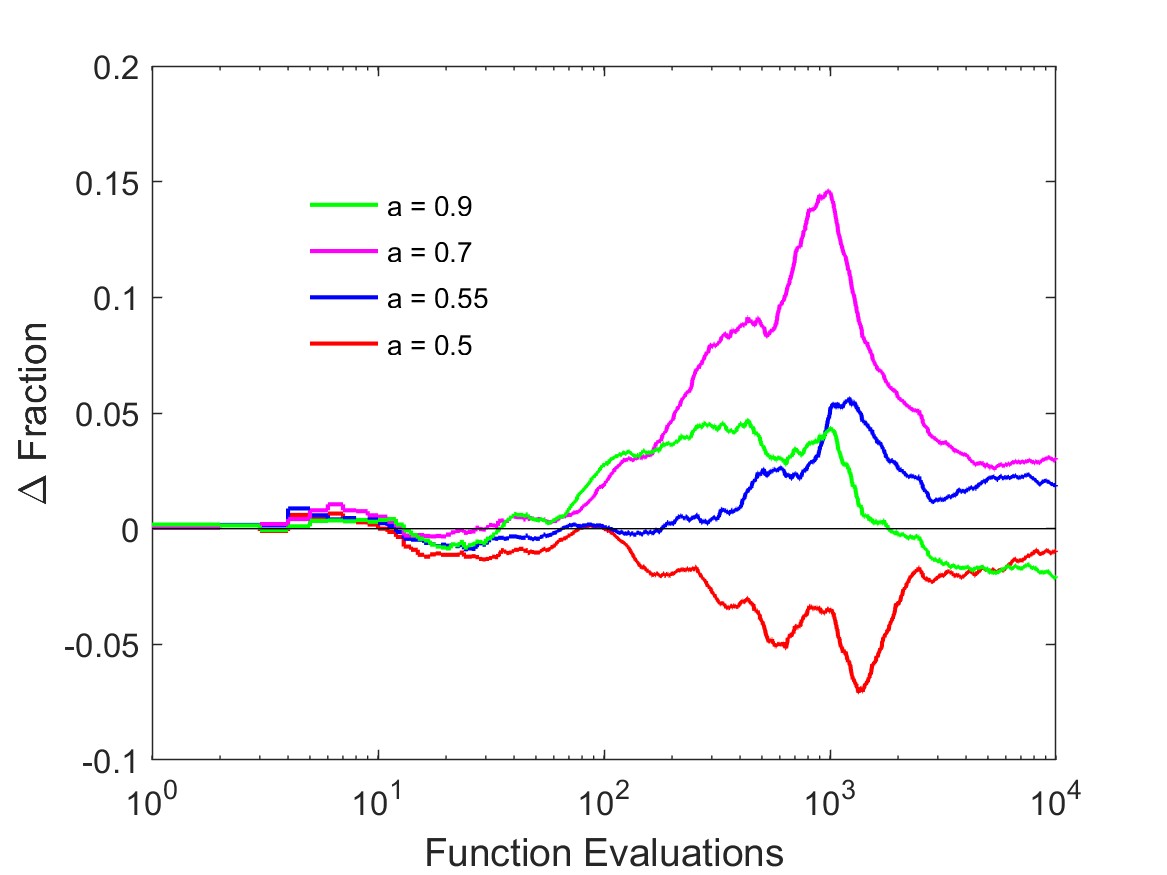}

 BBOB 1-14: \hspace{0.3cm} (b) Arcsine  \hspace{2.2cm} (e) Uniform \hspace{1.3cm}

\includegraphics[trim =7mm 3mm 10mm 8mm,clip, width=6.2cm, height=4.5cm]{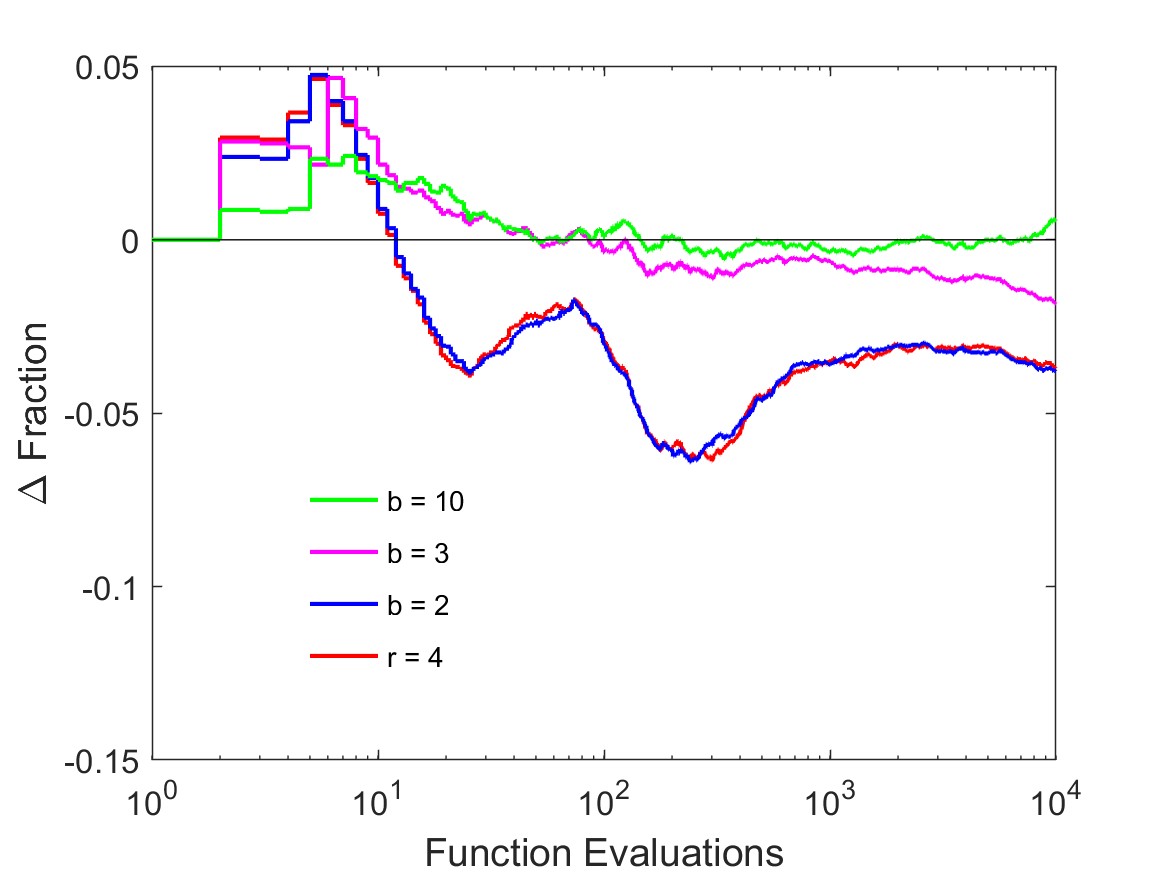}
\includegraphics[trim = 5mm 3mm 10mm 8mm,clip, width=6.2cm, height=4.5cm]{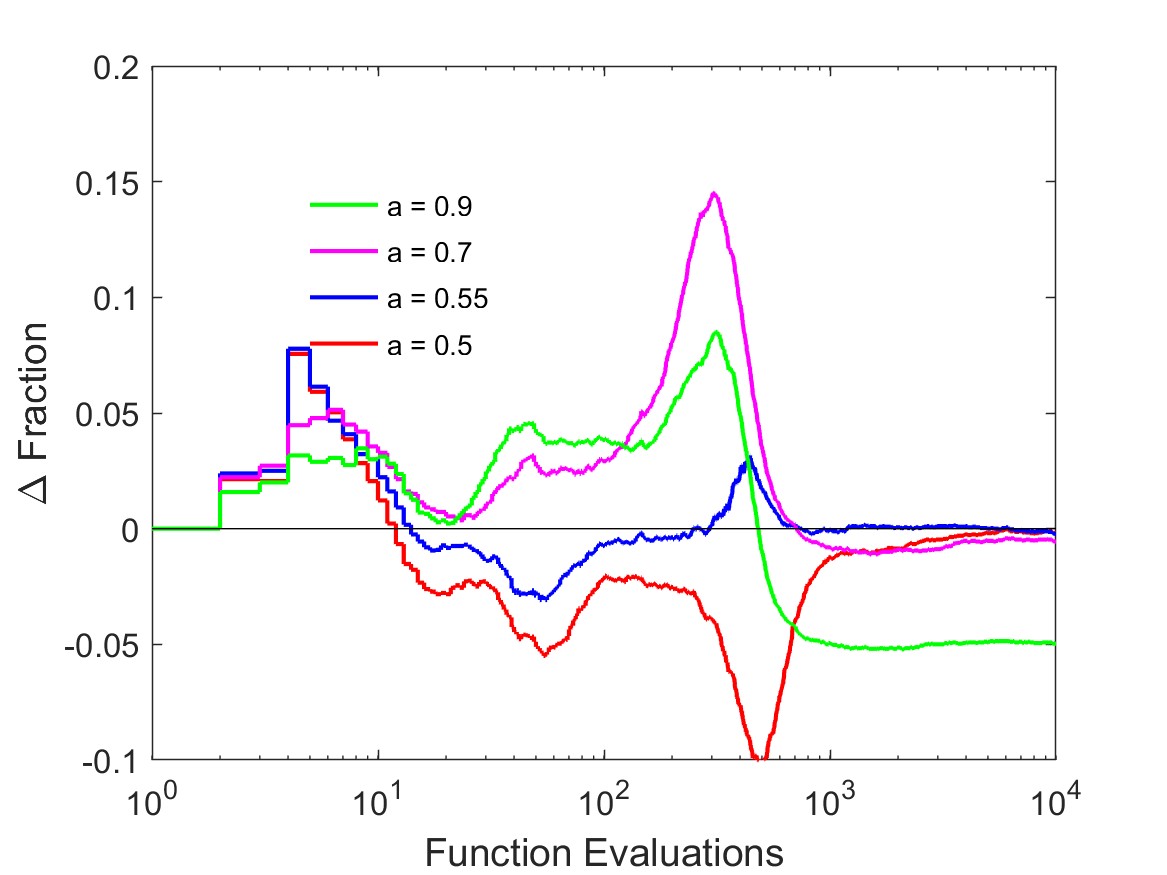}

BBOB 15-24: \hspace{0.2cm} (c) Arcsine  \hspace{2.2cm} (f) Uniform \hspace{1.3cm}
 
\caption{ $\Delta \: \text{Fraction} = \text{EAF}_{chao}-\text{EAF}_{rand}$. Search space dimension $D=50$.}
\label{fig_anytime_comp_50}

\end{figure}

Figs.~\ref{fig_anytime_comp_10} and \ref{fig_anytime_comp_50} show 
differences between the EAF curves for chaotic and random values: $\Delta \: \text{Fraction} = \text{EAF}_{chao}-\text{EAF}_{rand}$ for the search space dimensions $D=\{10,50\}$, and the whole set as well as the unimodal and multimodal subsets of the BBOB  benchmark suite. 
Results for \(D=30\), along with a detailed discussion, can be found in Sec. 4 and Fig. 4 of the main text. The main patterns observed for \(D=30\) are generally consistent across the other state-space dimensions,  for instance, chaos also here seems to be universally advantageous for small budgets and we have a deterioration in the performance of chaotic values for larger budgets. Moreover, for uniform distributions also peaks in performance difference between random and chaos occur for  intermediate budgets, and the peaks are slightly shifted for unimodal vs. multimodal BBOB functions. The main difference is that for arcsine distribution and $D=10$ the peaks appear for $a=0.9$, while for $D=30$ the map parameters $a=0.7$ and $a=0.9$ are close to each other, while for $D=50$ we get the highest peaks for $a=0.7$.

\clearpage

\subsection*{S4: AUC for budget range $[20,200]$}

\begin{table}[htbp]
\centering
\caption{Area under the EAF curves (AUC) quantification for the whole numerical budget $[2,200]$:  means (M),  and lower and upper 95\% bootstrap confidence intervals (CI) using 5000 bootstrap samples over 15 PSO runs within each of the $24 \times 3=72$ BBOB function–instance combinations. Upper half: arcsine distributed values by the Schröder map, Eq.~\eqref{eq:schr} and $\mathcal{B}(z,0.5,0.5)$. Lower half: uniform distributed values by the asymmetric tent map, Eq.~\eqref{eq:atent} and $\mathcal{B}(z,1,1)$. Last lines labeled by R: random values. Bold print: best AUC for each $D$ and distribution.   }
\label{tab:boots_2}
\scalebox{0.75}{
\begin{tabular}{c| cc | cc |cc}
\toprule
 & \multicolumn{2}{c}{$D=10$} & \multicolumn{2}{c}{$D=30$} & \multicolumn{2}{c}{$D=50$} \\
\cmidrule(lr){2-3} \cmidrule(lr){4-5} \cmidrule(lr){6-7}
r/b & M & CI & M & CI & M & CI\\
 \midrule
4 &  0.3773    &  [0.3179,    0.4397]  &    0.1701    &  [0.1242,    0.2276]  &    0.1173    &  [0.0786,    0.1655]\\
2 &   0.3805    &  [0.3212,    0.4465]  &    0.1704    &  [0.1244,    0.2274]  &    0.1176    &  [0.0789,    0.1674]\\
3 &   0.4191    &  [0.3589,    0.4821]  &    0.1868    &  [0.1387,    0.2430]  &    0.1387    &  [0.0964,    0.1910]\\
4 &   0.4362    &  [0.3750,    0.5015]  &    0.1908    &  [0.1437,    0.2508]  &    0.1422    &  [0.1009,    0.1953]\\
6 &   0.4383    &  [0.3757,    0.5006]  &    0.1936    &  [0.1442,    0.2508]  &    0.1436    &  [0.0999,    0.1975]\\
8 &   0.4364    &  [0.3752,    0.5021]  &    0.1910    &  [0.1408,    0.2478]  &    0.1432    &  [0.1010,    0.1961]\\
10 &   0.4403    &  [0.3805,    0.5076]  &    0.1913    &  [0.1410,    0.2466]  &    0.1423    &  [0.1014,    0.1950]\\
R  &  0.4427    &  [0.3806,    0.5056]  &    0.1902    &  [0.1405,    0.2471]  &    0.1408    &  [0.0981,    0.1940]\\
\midrule
a & M & CI & M & CI & M & CI\\
 \midrule
 0.5 &  0.5194  &  [0.4580,   0.5793]  &  0.2380   &  [0.1865,    0.2958]   &  0.1561     &  [0.1146,    0.2060]\\
 0.55 &  0.5592 &   [0.4999,    0.6188] &   0.2657    &  [0.2119,    0.3225]  &   0.1677     &  [0.1251,    0.2171]\\
 0.6 &  0.5826  &  [0.5218,    0.6413]  &  0.2943    &  [0.2380,    0.3514]   &  0.1776     &  [0.1350,    0.2301]\\
 0.65&   0.5943 &    [0.5357,    0.6529] &   0.3171    &  [0.2618,    0.3789]  &   0.1883     &  [0.1439,    0.2396]\\
 0.7 &  0.6151 &   [0.5569,    0.6739]  &  0.3378    &  [0.2808,    0.3971]   &  0.1974     &  [0.1527,    0.2495]\\
 0.8 &  0.6250 &   [0.5642,    0.6804]  &  0.3474    &  [0.2892,    0.4083]   &  0.1971     &  [0.1526,    0.2496]\\
 0.9 &  0.6258 &   [0.5691,    0.6817]  &  0.3440    &  [0.2890,    0.4081]   &  0.1973     &  [0.1534,    0.2520]\\
 R  & 0.5582 &   [0.4955,    0.6189]  &  0.2712    &  [0.2160,    0.3318]   &  0.1720     &  [0.1286,    0.2206]\\
\bottomrule
\end{tabular}}
\end{table}

\end{document}